\documentclass[sigconf]{acmart}

\AtBeginDocument{%
  \providecommand\BibTeX{{%
    \normalfont B\kern-0.5em{\scshape i\kern-0.25em b}\kern-0.8em\TeX}}}

\copyrightyear{2023} 
\acmYear{2023} 
\setcopyright{acmlicensed}\acmConference[WWW '23]{Proceedings of the ACM
Web Conference 2023}{April 30-May 4, 2023}{Austin, TX, USA}
\acmBooktitle{Proceedings of the ACM Web Conference 2023 (WWW '23), April
30-May 4, 2023, Austin, TX, USA}
\acmPrice{15.00}
\acmDOI{10.1145/3543507.3583455}
\acmISBN{978-1-4503-9416-1/23/04}

\acmSubmissionID{7089}

\usepackage{subfigure}
\usepackage{multirow}
\usepackage{bm}

\begin{document}

\title{HGWaveNet: A Hyperbolic Graph Neural Network for Temporal Link Prediction}

\author{Qijie Bai}
\affiliation{%
  \institution{College of CS, TJ Key Lab of NDST \\ Nankai University}
  \city{Tianjin}
  \country{China}}
\email{qijie.bai@mail.nankai.edu.cn}

\author{Changli Nie}
\affiliation{%
  \institution{College of CS, TJ Key Lab of NDST \\ Nankai University}
  \city{Tianjin}
  \country{China}}
\email{nie_cl@mail.nankai.edu.cn}

\author{Haiwei Zhang}
\authornote{Corresponding author.}
\affiliation{%
  \institution{College of CS, TJ Key Lab of NDST \\ Nankai University}
  \city{Tianjin}
  \country{China}}
\email{zhhaiwei@nankai.edu.cn}

\author{Dongming Zhao}
\affiliation{%
  \institution{China Mobile Communication Group Tianjin Co., Ltd}
  \city{Tianjin}
  \country{China}}
\email{waitman_840602@163.com}

\author{Xiaojie Yuan}
\affiliation{%
  \institution{College of CS, TJ Key Lab of NDST \\ Nankai University}
  \city{Tianjin}
  \country{China}}
\email{yuanxj@nankai.edu.cn}

\renewcommand{\shortauthors}{Qijie Bai, Changli Nie, Haiwei Zhang, Dongming Zhao and Xiaojie Yuan}

\begin{abstract}
Temporal link prediction, aiming to predict future edges between paired nodes in a dynamic graph, is of vital importance in diverse applications.
%
However, existing methods are mainly built upon uniform Euclidean space, which has been found to be conflict with the power-law distributions of real-world graphs and unable to represent the hierarchical connections between nodes effectively.
With respect to the special data characteristic, hyperbolic geometry offers an ideal alternative due to its exponential expansion property.
In this paper, we propose HGWaveNet, a novel hyperbolic graph neural network that fully exploits the fitness between hyperbolic spaces and data distributions for temporal link prediction.
Specifically, we design two key modules to learn the spatial topological structures and temporal evolutionary information separately.
On the one hand, a hyperbolic diffusion graph convolution (HDGC) module effectively aggregates information from a wider range of neighbors.
On the other hand, the internal order of causal correlation between historical states is captured by hyperbolic dilated causal convolution (HDCC) modules.
The whole model is built upon the hyperbolic spaces to preserve the hierarchical structural information in the entire data flow.
%
To prove the superiority of HGWaveNet, extensive experiments are conducted on six real-world graph datasets and the results show a relative improvement by up to 6.67\% on AUC for temporal link prediction over SOTA methods.
\end{abstract}



\maketitle

\section{Introduction}  \label{sec::intro}

Dynamic graphs, derived from the general graphs by adding an additional temporal dimension, have attracted a lot of attention in recent years~\cite{holmeTemporalNetworks2012}.
It has become a general practice to abstract real-world complex
systems (e.g. traffic systems~\cite{zhaoTGCNTemporalGraph2020}, social networks~\cite{yangRelationLearningSocial2020a} and e-commerce platforms~\cite{jiWhoYouWould2021}) to this newly data structure due to its powerful ability for modeling the temporal interactions between entities.
Temporal link prediction, aiming to forecast future relationships between nodes, is of vital importance for understanding the evolution of dynamic graphs~\cite{yangFewshotLinkPrediction2022}.

According to the expression of temporal information, dynamic graphs can be divided into two categories: continuous dynamic graphs and discrete dynamic graphs~\cite{zakiComprehensiveSurveyDynamic2016}.
%
Continuous dynamic graphs, also called graph streams, can be viewed as groups of edges ordered by time and each edge is associated with a timestamp. 
Recording topological structures of a dynamic graph at constant time intervals as snapshots, the list of snapshots is defined as a discrete dynamic graph.
In comparison, continuous dynamic graphs keep all temporal information~\cite{kempeConnectivityInferenceProblems}, but discrete dynamic graphs are more computationally efficient benefiting from the coarse-grained updating frequency~\cite{naokiGuideTemporalNetworks}.
This paper studies temporal link prediction on discrete dynamic graphs.
%


Existing studies have put forward some approaches to dynamic graph embedding and temporal link prediction. 
Most of them focus on two main contents: the \textbf{spatial topological structure} and the \textbf{temporal evolutionary information} of dynamic graphs.
For instance, CTDNE~\cite{nguyenContinuousTimeDynamicNetwork2018} runs temporal random walks to capture both spatial and temporal information on graphs.
TDGNN~\cite{quContinuousTimeLinkPrediction2020} introduces a temporal aggregator for GNNs to aggregate historical information of neighbor nodes and edges. 
Hawkes process~\cite{zuoEmbeddingTemporalNetwork2018, huangTemporalHeterogeneousInformation2021} is also applied to simulate the evolution of graphs. 
Furthermore, DyRep~\cite{trivediDYREPLEARNINGREPRESENTATIONS2019} leverages recurrent neural networks to update node representations over time.
TGAT~\cite{xuInductiveRepresentationLearning2020} uses the multi-head attention mechanism~\cite{vaswaniAttentionAllYou} and a sampling strategy that fixes the number of neighbor nodes involved in message-propagation to learn both spatial and temporal information.

\begin{figure}[t]
    \centering
    \includegraphics[width=0.9\linewidth]{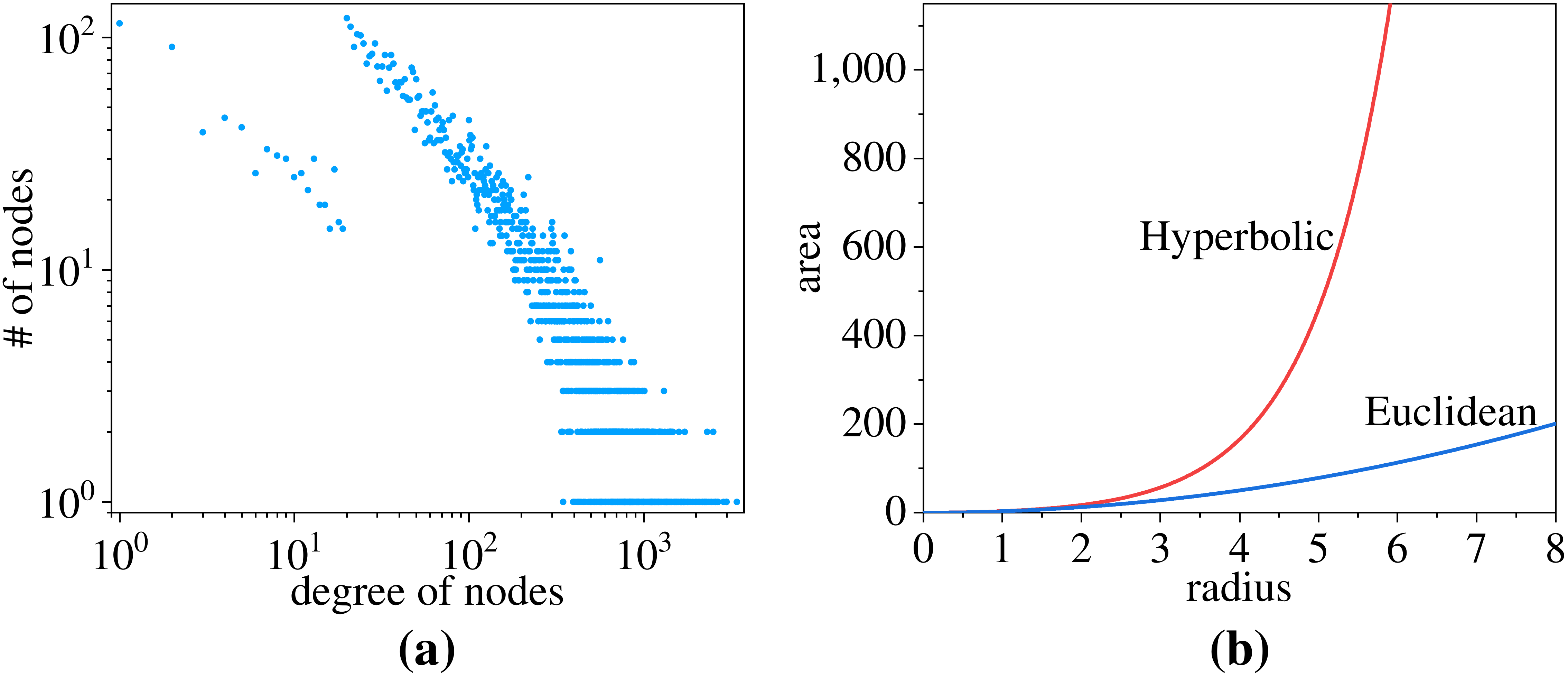}
    \caption{(a) The degree distribution of real-world network MovieLens. Coordinate axes are logarithmic. (b) Areas of circles on hyperbolic spaces (curvature $c=-1$) and Euclidean spaces, which shows that the hyperbolic spaces expand exponentially while Euclidean spaces expand polynomially.}
    \Description{Degree distributions of MovieLens and .}
    \label{fig::intro:distributions}
\end{figure}

These methods are all built upon Euclidean spaces. 
However, recent works~\cite{bronsteinGeometricDeepLearning2017, krioukovHyperbolicGeometryComplex2010} have noticed that most real-world graph data, such as social networks, always exhibit implicit hierarchical structures and power-law distributions (as shown in Figure~\ref{fig::intro:distributions}(a)) rather than uniform grid structures which fit Euclidean spaces best.
The mismatches between data distributions and space geometries severely limit the performances of Euclidean models~\cite{nickelPoincareEmbeddingsLearning, chamiHyperbolicGraphConvolutional}.
%


Hyperbolic spaces are those of constant negative curvatures, and are found to keep great representation capacity for data with hierarchical structures and power-law distributions due to the exponential expansion property (as shown in Figure~\ref{fig::intro:distributions}(b))~\cite{krioukovHyperbolicGeometryComplex2010, pengHyperbolicDeepNeural2021}.
In the past few years, researchers have made some progress in representing hierarchical data in hyperbolic spaces~\cite{nickelPoincareEmbeddingsLearning, liuHyperbolicGraphNeural, chamiHyperbolicGraphConvolutional, zhangHyperbolicGraphAttention2021, wangHyperbolicHeterogeneousInformation2019, nickelLearningContinuousHierarchiesa, salaRepresentationTradeoffsHyperbolic, yangDiscretetimeTemporalNetwork2021}.
HTGN~\cite{yangDiscretetimeTemporalNetwork2021} attempts to embed dynamic graphs into hyperbolic geometry and achieves state-of-the-art performance. 
It adopts hyperbolic GCNs to capture spatial features and a hyperbolic temporal contextual attention module to extract the historical information.
However, this method faces two major shortcomings. 
First, for some large graphs with long paths, the message propagation mechanism of GCNs is not effective enough because only direct neighbor nodes can be calculated in each step.
Second, the temporal contextual attention module cannot handle the internal order of causal correlation in historical data and thus loses valid information for temporal evolutionary process.

To address the two aforementioned shortcomings of HTGN, in this paper, we propose a novel hyperbolic graph neural network model named HGWaveNet for temporal link prediction. 
In HGWaveNet, we first project the nodes into hyperbolic spaces.
Then in the aspect of spatial topology, a hyperbolic diffusion graph convolution (HDGC) module is designed to learn node representations of each snapshot effectively from both direct neighbors and indirectly connected nodes.
For temporal information, recurrent neural networks always emphasize short time memory while ignoring that of long time due to the time monotonic assumption~\cite{choLearningPhraseRepresentations2014}, and the attention mechanism ignores the causal orders in temporality.
Inspired by WaveNet~\cite{oordWaveNetGenerativeModel2016} and Graph WaveNet~\cite{wuGraphWaveNetDeep2019}, we present hyperbolic dilated causal convolution (HDCC) modules to obtain hidden cumulative states of nodes by aggregating historical information and capturing temporal dependencies. 
In the training process of each snapshot, the hidden cumulative states and spatial-based node representations are fed into a hyperbolic gated recurrent unit (HGRU).
The outputs of HGRU are regarded as the integration of spatial topological structures and temporal evolutionary information, and are utilized for final temporal link prediction. 
A hyperbolic temporal consistency (HTC)~\cite{yangDiscretetimeTemporalNetwork2021} component is also leveraged to ensure stability for tracking the evolution of graphs.

To conclude, the main contributions of this paper are as follows:
\begin{itemize}
\item We propose a novel hyperbolic graph neural network model named HGWaveNet for temporal link prediction, which learns both spatial hierarchical structures and temporal causal correlation of dynamic graphs.
\item With regard to spatial topological structures, we design a hyperbolic diffusion graph convolution (HDGC) module to fit the power-law distributions of data and aggregate information from a wider range of nodes with greater effectiveness.
\item For temporal evolution, we present hyperbolic dilated causal convolution (HDCC) modules to capture the internal causality between snapshots, and a hyperbolic temporal consistency (HTC) component is applied to remain stable when learning the evolution of graphs.
\item We conduct extensive experiments on diverse real-world dynamic graphs. The results prove the superiority of HGWaveNet, as it has a relative improvement by up to 6.67\% in terms of AUC over state-of-the-art methods.
\end{itemize}

\section{Related Work}  \label{sec::related}

In this section, we systematically review the relevant works on temporal link prediction and hyperbolic graph representation learning.

\subsection{Temporal Link Prediction}  \label{sec::related:temporal}

Temporal link prediction on dynamic graphs has attracted increasing interests in the past few years.
Early methods usually use traditional algorithms or shallow neural architectures to represent structural and temporal information. 
For instance, CTDNE~\cite{nguyenContinuousTimeDynamicNetwork2018} captures the spatial and temporal information simultaneously by adding temporal constraints to random walk.
Another example of using temporal random walk is DynNode2vec~\cite{mahdaviDynnode2vecScalableDynamic2018}, which updates the sampled sequences incrementally at each snapshot rather than generating them anew.
DynamicTriad~\cite{zhouDynamicNetworkEmbedding} imposes the triad closure process and models the evolution of graphs by developing closed triads from open triads.
Furthermore, Change2vec~\cite{bianNetworkEmbeddingChange2019} improves this process and makes it applicable for dynamic heterogeneous graphs.
HTNE~\cite{zuoEmbeddingTemporalNetwork2018} integrates the Hawkes process into graph embedding to learn the influence of historical neighbors on current neighbors.

In contrast, graph neural network methods have recently received increasing attention. 
GCN~\cite{kipfSemiSupervisedClassificationGraph2017} provides an excellent node embedding pattern for general tasks on graphs, and most of later models take GCN or its variants, such as GraphSAGE~\cite{hamiltonInductiveRepresentationLearning} and GAT~\cite{velickovicGraphAttentionNetworks2018}, as basic modules for learning topological structures.
GCRN~\cite{seoStructuredSequenceModeling2016} feeds node representations learned from GCNs into a modified LSTM to obtain the temporal information. 
Similar ideas are explored in EvolveGCN~\cite{parejaEvolveGCNEvolvingGraph2020}, E-LSTM-D~\cite{chenELSTMDDeepLearning2021} and NTF~\cite{wuNeuralTensorFactorization2019}.
The main difference between EvolveGCN and E-LSTM-D is that EvolveGCN could be considered as a combination of GCN and RNN while E-LSTM-D uses LSTM together with an encoder-decoder architecture.
NTF takes a reverse order that characterizes the temporal interactions with LSTM before adopting the MLP for non-linearities between different latent factors, and sufficiently learns the evolving characteristics of graphs.
To better blend the spatial and temporal node features, DySAT~\cite{sankarDySATDeepNeural2020} proposes applying multi-head self-attention.
TGN~\cite{rossiTemporalGraphNetworks2020a} leverages memory modules with GCN operators and significantly increases the computational efficiency. 
TNS~\cite{wangTimeAwareNeighborSampling2021} provides an adaptive receptive neighborhood for each node at any time.
VRGNN~\cite{hajiramezanaliVariationalGraphRecurrent} models the uncertainty of node embeddings by regarding each node in each snapshot as a distribution.

However, the prevailing methods are built upon Euclidean spaces, which are not isometric with the power-law distributions of real-world graphs, and may cause distortion with the graph scale grows.

\subsection{Hyperbolic Graph Representation Learning}  \label{sec::related:hyperbolic}

Representation learning in hyperbolic spaces has been noticed due to their fitness to the hierarchical structures of real-world data.
The significant performance advantages shown by the shallow Poincaré~\cite{nickelPoincareEmbeddingsLearning} and Lorentz~\cite{nickelLearningContinuousHierarchiesa} models spark more attempts to this issue. 
Integrated with graph neural networks, HGNN~\cite{liuHyperbolicGraphNeural} and HGCN~\cite{chamiHyperbolicGraphConvolutional} are designed for graph classification and node embedding tasks separately.
HAT~\cite{zhangHyperbolicGraphAttention2021} exploits attention mechanism for hyperbolic node information propagation and aggregation.
LGCN~\cite{zhangLorentzianGraphConvolutional2021} builds the graph operations of hyperbolic GCNs with Lorentzian version and rigorously guarantees that the learned node features follow the hyperbolic geometry.
The above hyperbolic GNN-based models all adopt the bi-directional transition between a hyperbolic space and corresponding tangent spaces, while a tangent space is the first-order approximation of the original space and may inevitably cause distortion.
To avoid distortion, H2H-GCN~\cite{daiHyperbolictoHyperbolicGraphConvolutional} develops a manifold-preserving graph convolution and directly works on hyperbolic manifolds.
For practical application situations, HGCF~\cite{sunHGCFHyperbolicGraph2021}, HRCF~\cite{yangHRCFEnhancingCollaborative2022} and HICF~\cite{yangHICFHyperbolicInformative2022} study the hyperbolic collaborative filtering for user-item recommendation systems.
Through hyperbolic graph learning, HyperStockGAT~\cite{sawhneyExploringScaleFreeNature2021} captures the scale-free spatial and temporal dependencies in stock prices, and achieves state-of-the-art stock forecasting performance.

HVGNN~\cite{sunHyperbolicVariationalGraph2021} and HTGN~\cite{yangDiscretetimeTemporalNetwork2021} fill the gap of hyperbolic models on dynamic graphs. 
HVGNN generates stochastic node representations of hyperbolic normal distributions via a hyperbolic graph variational autoencoder to represent the uncertainty of dynamics.
HTGN adopts a conventional model architecture that handles the spatial and temporal information with HGCNs and contextual attention modules separately, however, ignores the causal order in the graph evolution.
To further improve the hyperbolic graph models especially on temporal link prediction problem, we propose our HGWaveNet in terms of discrete dynamic graphs.

\section{Preliminaries}  \label{sec::preli}

In this section, we first give the formalized definitions of discrete dynamic graphs and temporal link prediction. 
Then some critical fundamentals about hyperbolic geometry are introduced.

\subsection{Problem Definition}  \label{sec::preli:problem}

This paper discusses temporal link prediction on discrete dynamic graphs. 
Following ~\cite{khoshraftarSurveyGraphRepresentation2022}, we define discrete dynamic graphs as:

\begin{definition}[\textbf{Discrete Dynamic Graphs}]  \label{def::preli:discrete}
\emph{
In discrete dynamic graph modeling, dynamic graphs can be viewed as a sequence of snapshots sampled from the original evolving process at consecutive time points.
Formally, discrete dynamic graphs are represented as $\mathcal{G}=(\mathcal{G}_0, \mathcal{G}_1, \cdots, \mathcal{G}_T)$, in which $\mathcal{G}_t$ is a snapshot at timestamp $t$. 
The time granularity for snapshot divisions could be hours, days, months or even years depending on specific datasets and applications.
}
\end{definition}

Based on Definition~\ref{def::preli:discrete}, temporal link prediction is described as:

\begin{definition}[\textbf{Temporal Link Prediction}]  \label{def::preli:temporal}
\emph{
The aim of temporal link prediction is to predict the links appeared in the snapshots after timestamp $t$ based on the observed snapshots before timestamp $t$.
Formally, the model takes $(\mathcal{G}_0, \mathcal{G}_1, \cdots, \mathcal{G}_{t-1})$ as input in the training process, and then makes predictions on $(\mathcal{G}_t, \mathcal{G}_{t+1}, \cdots, \mathcal{G}_T)$.
}
\end{definition}

\subsection{Hyperbolic Geometry of the Poincaré Ball}  \label{sec::preli:hyperbolic}

The $n$-dimensional hyperbolic space $(\mathbb{H}_c^n, g^{c, \mathbb{H}})$ is the unique simply connected $n$-dimensional complete Riemannian manifold $\mathbb{H}_c^n$ with a constant negative curvature $-c (c > 0)$, and $g^{c, \mathbb{H}}$ is the Riemannian metric.
The tangent space $\mathcal{T}_{\textbf{x}} \mathbb{H}_c^n$ is a Euclidean, local, first-order approximation of $\mathbb{H}_c^n$ around the point $\textbf{x} \in \mathbb{H}_c^n$. 
Similar to ~\cite{nickelPoincareEmbeddingsLearning} and ~\cite{ganeaHyperbolicNeuralNetworks}, we construct our method based on the Poincaré ball, one of the most widely used isometric models of hyperbolic spaces. 
Corresponding to $(\mathbb{H}_c^n, g^{c, \mathbb{H}})$, the Poincaré ball $(\mathbb{B}_c^n, g^{c, \mathbb{B}})$ is defined as
\begin{equation}
\begin{split}
    & \mathbb{B}_c^n := \left\{ \textbf{x} \in \mathbb{R}^n: c \| \textbf{x} \|^2 < 1 \right\}, \\
    & g^{c, \mathbb{B}}_{\textbf{x}} := \left( \lambda_{\textbf{x}}^c \right)^2 g^{E}, \quad \lambda_{\textbf{x}}^c := \frac{2}{1 - c \| \textbf{x} \|^2},
    \label{eq::preli:poincare}
\end{split}
\end{equation}
where $g^{E} = \textbf{I}_n$ is the Euclidean metric tensor. 

\begin{figure}[t]
    \centering
    \includegraphics[width=0.4\linewidth]{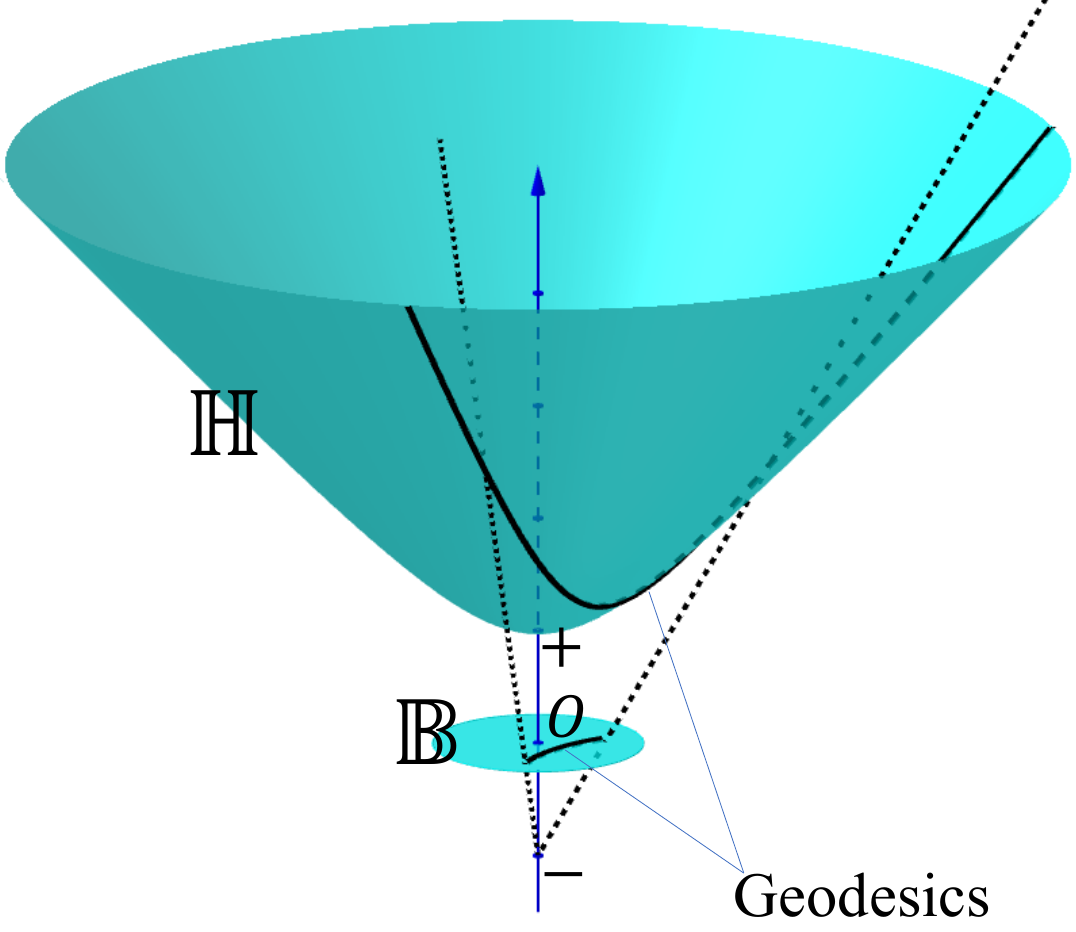}
    \caption{A hyperbolic manifold $\mathbb{H}$ and the corresponding Poincaré ball model $\mathbb{B}$. The distances on the manifolds are measured along geodesics.}
    \Description{The Poincaré ball model.}
    \label{fig::preli:poincare}
\end{figure}

The Poincaré ball manifold $\mathbb{B}_c^n$ is an open ball of radius $1 / \sqrt{c}$ (see Figure~\ref{fig::preli:poincare}).
The induced distance between two points $\textbf{x}, \textbf{y} \in \mathbb{B}_c^n$ is measured along a geodesic and given by\footnote{$- \textbf{x} \oplus_c \textbf{y}$ should be read as $(- \textbf{x}) \oplus_c \textbf{y}$ rather than $- (\textbf{x} \oplus_c \textbf{y})$.}
\begin{equation}
    d_c \left( \textbf{x}, \textbf{y} \right) = \frac{2}{\sqrt{c}} \cdot \tanh^{-1} \left( \sqrt{c} \| - \textbf{x} \oplus_c \textbf{y} \| \right),
    \label{eq::preli:distance}
\end{equation}
in which the Möbius addition $\oplus_c$ in $\mathbb{B}_c^n$ is defined as
\begin{equation}
    \textbf{x} \oplus_c \textbf{y} := \frac{\left( 1 + 2 c \langle \textbf{x}, \textbf{y} \rangle + c \| \textbf{y} \|^2 \right) \textbf{x} + \left( 1 - c \| \textbf{x} \|^2 \right) \textbf{y}}{1 + 2 c \langle \textbf{x}, \textbf{y} \rangle + c^2 \| \textbf{x} \|^2 \| \textbf{y} \|^2}.
    \label{eq::preli:addition}
\end{equation}

To map points between hyperbolic spaces and tangent spaces, exponential and logarithmic maps are given~\cite{ganeaHyperbolicNeuralNetworks}.
For $\textbf{x}, \textbf{y} \in \mathbb{B}_c^n, \textbf{v} \in \mathcal{T}_{\textbf{x}} \mathbb{B}_c^n, \textbf{x} \neq \textbf{y}$ and $\textbf{v} \neq \textbf{0}$, the exponential map $\exp_{\textbf{x}}^c: \mathcal{T}_{\textbf{x}} \mathbb{B}_c^n \rightarrow \mathbb{B}_c^n$ is
\begin{equation}
    \exp_{\textbf{x}}^c (\textbf{v}) = \textbf{x} \oplus_c \left( \tanh \left( \frac{\sqrt{c} \lambda_{\textbf{x}}^c \| \textbf{v} \|}{2} \right) \frac{\textbf{v}}{\sqrt{c} \| \textbf{v} \|} \right),
    \label{eq::preli:expmap}
\end{equation}
and the logarithmic map $\log_{\textbf{x}}^c: \mathbb{B}_c^n \rightarrow \mathcal{T}_{\textbf{x}} \mathbb{B}_c^n$ is
\begin{equation}
    \log_{\textbf{x}}^c (\textbf{y}) = d_c \left( \textbf{x}, \textbf{y} \right) \frac{- \textbf{x} \oplus_c \textbf{y}}{\lambda_{\textbf{x}}^c \| - \textbf{x} \oplus_c \textbf{y} \|},
    \label{eq::preli:logmap}
\end{equation}
where $\lambda_{\textbf{x}}^c$ and $d_c \left( \textbf{x}, \textbf{y} \right)$ are the same as in Equations~(\ref{eq::preli:poincare}) and ~(\ref{eq::preli:distance}).
In our method, we use the origin point $\textbf{o}$ as the reference point $\textbf{x}$ to balance the errors in diverse directions.

\section{Methodology}  \label{sec::method}

This section describes our proposed model HGWaveNet. 
First, we elaborate on the details of two core components: hyperbolic diffusion graph convolution (HDGC) and hyperbolic dilated causal convolution (HDCC). 
Then other key modules contributing to HGWaveNet are introduced, including gated HDCC, hyperbolic gated recurrent unit (HGRU), hyperbolic temporal consistency (HTC) and Fermi-Dirac decoder. 
Finally, we summarize the overall framework of HGWaveNet and analyse the time complexity.

\subsection{Hyperbolic Diffusion Graph Convolution}  \label{sec::method:diffusion}

Hyperbolic graph convolutional neural networks (HGCNs~\cite{chamiHyperbolicGraphConvolutional}) are built analogous to traditional GNNs. 
A typical HGCN layer consists of three key parts: hyperbolic feature transform
\begin{equation}
    \textbf{h}_i^{l, \mathbb{B}} = \textbf{W}^l \otimes_{c^{l-1}} \textbf{x}_i^{l-1, \mathbb{B}} \oplus_{c^{l-1}} \textbf{b}^l,
    \label{eq::method:linear}
\end{equation}
attention-based neighbor aggregation
\begin{equation}
\begin{split}
    \textbf{y}_i^{l, \mathbb{B}} &= \text{Agg}_{c^l} (\textbf{h}^{l, \mathbb{B}})_i \\
    &= \exp_{\textbf{o}}^{c^l} \left( \text{Att} \left( \text{Concat}_{j \in \mathcal{N}(i)} \left( \log_{\textbf{o}}^{c^{l-1}} \left( \textbf{h}_j^{l, \mathbb{B}} \right) \right) \right) \right), 
    \label{eq::method:aggregation}
\end{split}
\end{equation}
and hyperbolic activation
\begin{equation}
    \textbf{x}_i^{l, \mathbb{B}} = \exp_{\textbf{o}}^{c^{l}} \left( \sigma \left( \log_{\textbf{o}}^{c^l} \left( \textbf{y}_i^{l, \mathbb{B}} \right) \right) \right),
    \label{eq::method:activation}
\end{equation}
in which $\textbf{W}^l, \textbf{b}^l$ are trainable parameters and $\textbf{x}_i^{l, \mathbb{B}}$ is the representation of node $i$ at layer $l$ in manifold $\mathbb{B}_{c^l}^n$. Following \cite{yangDiscretetimeTemporalNetwork2021}, the matrix-vector multiplication $\otimes_c$ is defined as
\begin{equation}
    \textbf{M} \otimes_c \textbf{x} := \exp_{\textbf{o}}^c \left( \textbf{M} \log_{\textbf{o}}^c \left( \textbf{x} \right) \right),  \quad  \textbf{M} \in \mathbb{R}^{n \times n}, \textbf{x} \in \mathbb{B}_c^n.
    \label{eq::method:multiplication}
\end{equation}

However, the shallow HGCN can only aggregate information of direct neighbors and is not effective enough for large graphs with long paths. 
To overcome this shortcoming, we impose the diffusion process referring to ~\cite{liDiffusionConvolutionalRecurrent2018}.
Consider a random walk process with restart probability $\alpha$ on $\mathcal{G}$, and a state transition matrix $\textbf{P}$ (for most situations, $\textbf{P}$ is the normalized adjacent matrix). 
Such Markov process converges to a stationary distribution $\mathcal{P} \in \mathbb{R}^{n \times n}$ after many steps, the $i$-th row of which is the likelihood of diffusion from node $i$. 
The stationary distribution $\mathcal{P}$ can be calculated in the closed form~\cite{tengScalableAlgorithmsData2016}
\begin{equation}
    \mathcal{P} = \sum_{k=0}^{\infty} \alpha (1 - \alpha)^k \textbf{P}^k,
    \label{eq::method:stationary}
\end{equation}
where $k$ is the diffusion step. 
In practice, a finite $K$-step truncation of the diffusion process is adopted and separate trainable weight matrices are added to each step for specific objective tasks.
Then, the diffusion convolution layer on graphs is defined as
\begin{equation}
    \textbf{Z} = \sum_{k=0}^{K} \textbf{A}^k \textbf{X} \textbf{W}_k,
    \label{eq::method:diffusion_conv}
\end{equation}
in which $\textbf{A}$ is the bi-direct adjacent matrix, $\textbf{X}$ is the input node features, $\textbf{W}_k$ is the weight matrix for the $k$-th diffusion step, and $\textbf{Z}$ is the output node representations.

\begin{figure}[t]
    \centering
    \includegraphics[width=0.9\linewidth]{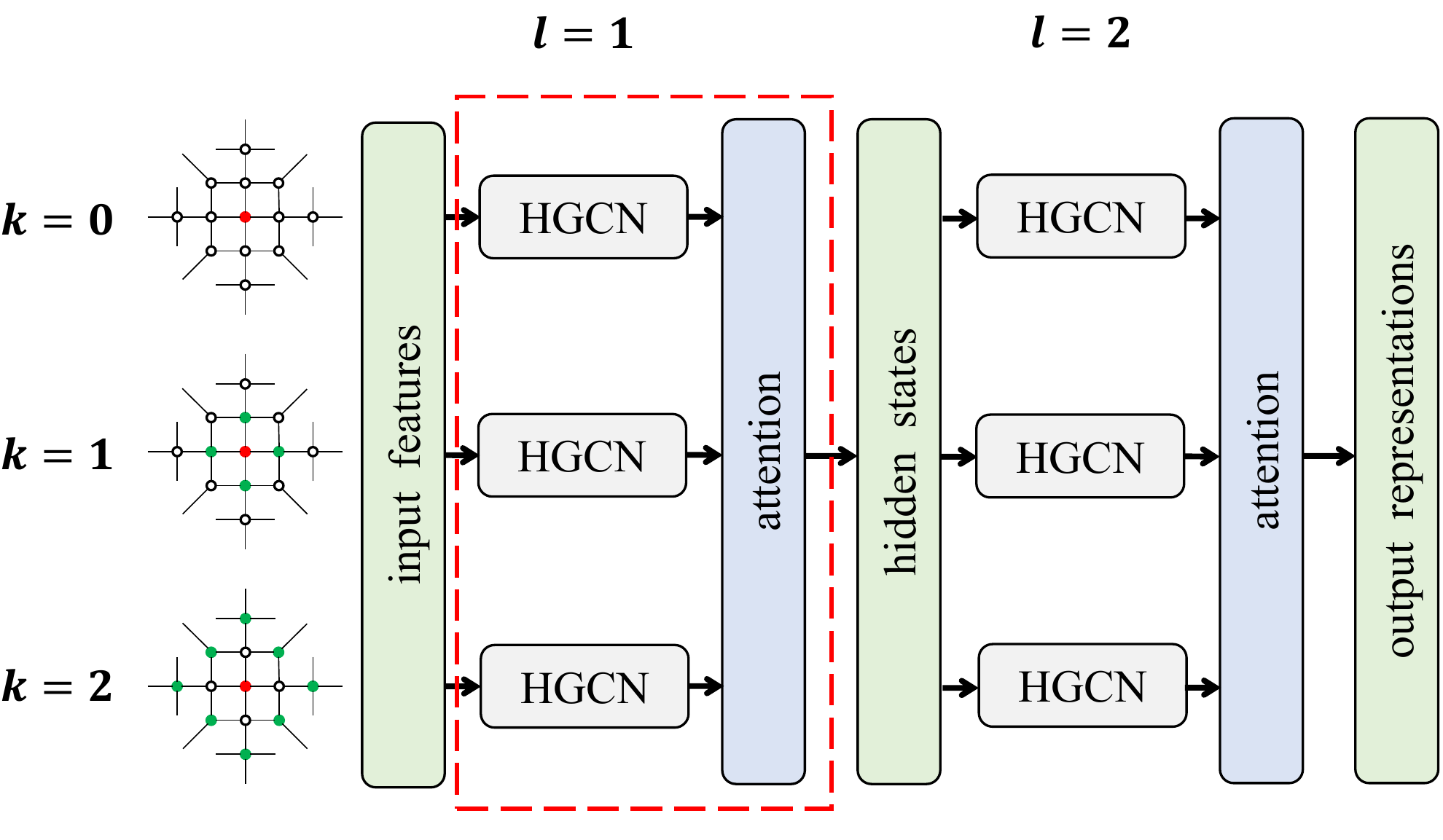}
    \caption{A hyperbolic diffusion graph convolution (HDGC) module with the number of layers $L = 2$ and the number of diffusion steps $K = 2$. The part in red dashed rectangle is a complete HDGC layer.}
    \Description{Hyperbolic diffusion graph convolution.}
    \label{fig::method:hdgc}
\end{figure}

On account of the sparsity of real-world data, we convert the graph convolution into equivalent spatial domains for efficiency. 
Do all above operations in the hyperbolic space, and replace the summation in Equation~(\ref{eq::method:diffusion_conv}) with an attention mechanism for better information aggregation.
Then, the $k$-th step of hyperbolic diffusion graph convolution at layer $l$ is
\begin{equation}
    \textbf{X}_k^l = \text{HGCN}_{c_k^l} \left( \textbf{A}^k, \textbf{X}^{l-1} \right),
    \label{eq::method:diffusion_conv_step}
\end{equation}
where $\text{HGCN}_{c^l} (\cdot)$ is the conjunctive form of Equations~(\ref{eq::method:linear}), ~(\ref{eq::method:aggregation}), and ~(\ref{eq::method:activation}) with the adjacent matrix and node features as inputs.
$\textbf{X}^{l}$ is calculated as
\begin{equation}
    \textbf{X}^l = \exp_{\textbf{o}}^{c^l} \left( \text{Att} \left( \text{Concat}_k \left( \log_{\textbf{o}}^{c_k^l} \left( \textbf{X}_k^l \right) \right) \right) \right).
    \label{eq::method:diffusion_aggregation}
\end{equation}
Hyperbolic diffusion graph convolution is finally constructed with the stack of layers defined by Equations~(\ref{eq::method:diffusion_aggregation}) and ~(\ref{eq::method:diffusion_conv_step}), as sketched in Figure~\ref{fig::method:hdgc}.

\subsection{Hyperbolic Dilated Causal Convolution}   \label{sec::method:dilated}

\begin{figure}[t]
    \centering
    \includegraphics[width=0.7\linewidth]{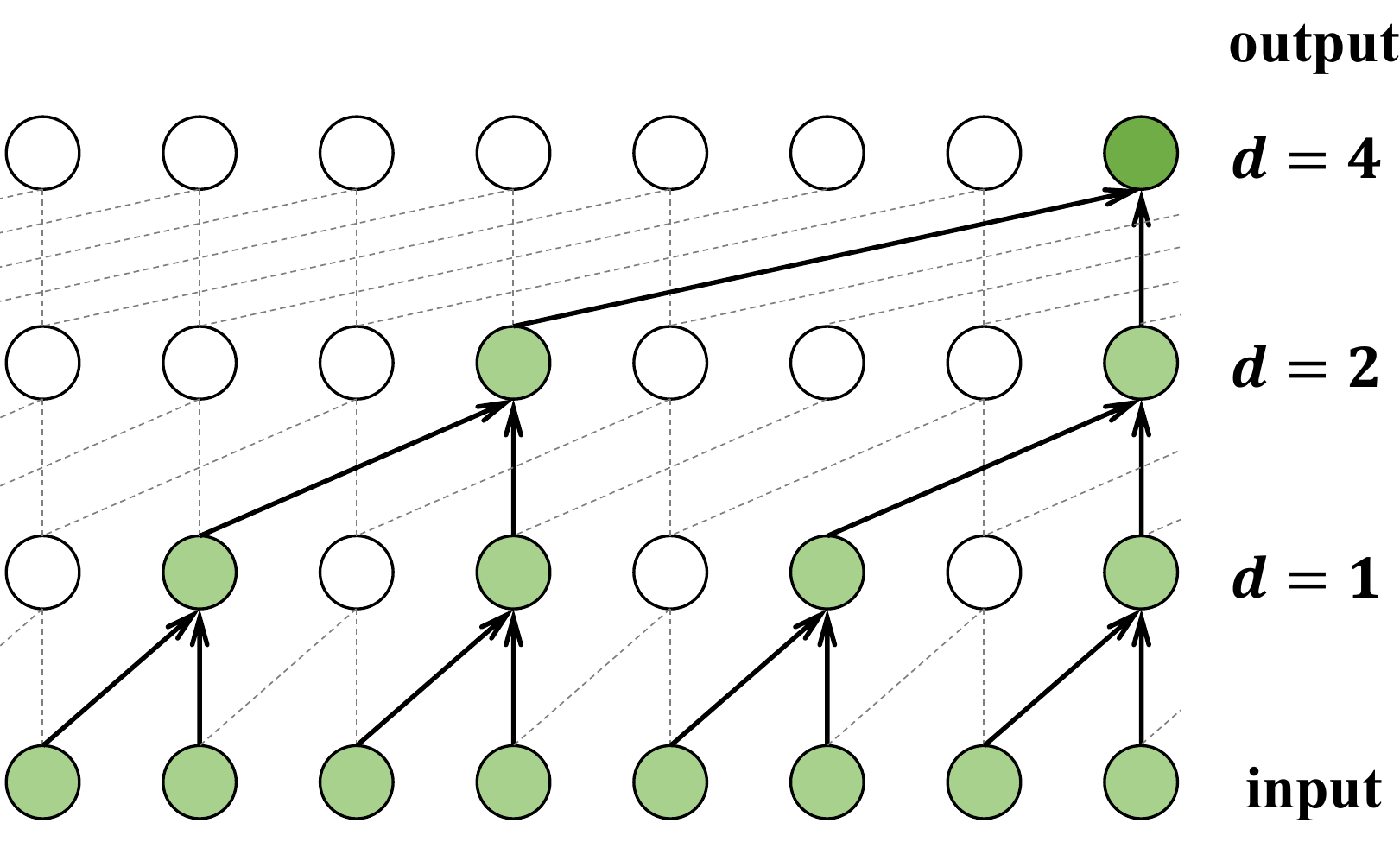}
    \caption{A stacked hyperbolic dilated causal convolution (HDCC) module with dilated depth $D = 3$ and kernel size $S = 2$. Each circle refers to one node at a timestamp. Under the exponentially large receptive field, each node preserves the temporal causal correlations with the past $S^D$ snapshots.}
    \Description{Hyperbolic dilated casual convolution.}
    \label{fig::method:dilated}
\end{figure}

Causal convolution refers to applying a convolution filter only to data at past timestamps~\cite{oordWaveNetGenerativeModel2016}, and guarantees that the order of data modeling is not violated, which means that the prediction $p( x_{t} | x_0, x_1, \cdots, x_{t-1} )$ emitted by the model at timestamp $t$ cannot depend on future data $x_t, x_{t+1}, \cdots, x_T$.
For 1-D temporal information, this operation can be implemented by shifting the output of a normal convolution with kernel size $S$ by $\lfloor S/2 \rfloor$ steps, but a sufficiently large receptive field requires stacking many layers.
To decrease the computational cost, dilated convolutions are adopted.
By skipping the inputs with a certain step $d$, a dilated convolution applies its kernel over a larger area than its length, and with the stack of dilated convolution layers, the receptive field expands exponentially while preserving high computational efficiency. 
Based on the above definitions of causal convolutions and dilated convolutions, the dilated causal convolution operation in manifold $\mathbb{B}_{c}^n$ can be formalized mathematically as
\begin{equation}
    (\textbf{Z}_i \odot \textbf{F})_t = \exp_{\textbf{o}}^c \left( \sum_{s=0}^{S-1} \textbf{F}_s \log_{\textbf{o}}^c \left( \textbf{Z}_{i, t - d \times s} \right) \right),
    \label{eq::method:dilated_causal_conv}
\end{equation}
where $d$ is the dilation step, $\textbf{Z}_{i, t} \in \mathbb{B}_{c}^n \ (t \leq T)$ is the representation of node $i$ at snapshot $\mathcal{G}_t$ and $\textbf{F} \in \mathbb{R}^{S \times n}$ denotes the kernels for all dimensions (also called channels).
A stacked hyperbolic dilated causal convolution module is shown in Figure~\ref{fig::method:dilated}.

Compared to attention-based historical information aggregation~\cite{yangDiscretetimeTemporalNetwork2021}, HDCC can learn the internal order of causal correlation in temporality. 
Meanwhile, compared to RNNs, HDCC allows parallel computation and alleviates vanishing or exploding gradient problems due to its non-recursion\cite{wuGraphWaveNetDeep2019}.
Considering the superiority of HDCC on sequence problems, we leverage this module in handling the temporal evolutionary information of graphs.

\subsection{Other Key Modules in HGWaveNet}   \label{sec::method:key_modules}

Except for HDGC and HDCC, some other modules contribute greatly to our proposed HGWaveNet. 
Based on HDCC, we further design gated HDCC to preserve the valid historical information. 
Hyperbolic gated recurrent unit (HGRU)~\cite{yangDiscretetimeTemporalNetwork2021} is able to efficiently learn the latest node representations in Poincaré ball from historical states and current spatial characteristics. 
Additionally, as defined in ~\cite{yangDiscretetimeTemporalNetwork2021}, hyperbolic temporal consistency (HTC) module provides a similarity constraint in temporality to ensure stability during graph evolution. 
Finally, a Fermi-Dirac decoder~\cite{krioukovHyperbolicGeometryComplex2010} is used to compute the probability scores for edge reconstruction.

\subsubsection{Gated HDCC}   \label{sec::method:key_modules:gated_hdcc}

As shown in the left part of Figure ~\ref{fig::method:hgwavenet}, we adopt a simple gating mechanism on the outputs of HDCCs. 
What is noteworthy is the use of logarithmic map before the Euclidean activation functions and exponential map after gating.
The formulation is described as
\begin{equation}
    \textbf{H}_{i, t} = \exp_{\textbf{o}}^c \left( \tanh \left( \log_{\textbf{o}}^c \left( \left( \textbf{Z}_i \odot \textbf{F}_1 \right)_t \right) \right) \cdot \sigma \left( \log_{\textbf{o}}^c \left( \left( \textbf{Z}_i \odot \textbf{F}_2 \right)_t \right) \right) \right),
    \label{eq::method:gated_hdcc}
\end{equation}
where $\textbf{F}_1$ and $\textbf{F}_2$ are different convolution kernels for two HDCCs, $\textbf{H}_{i, t} \in \mathbb{B}_{c}^n \ (t \leq T)$ is the historical hidden state of node $i$ at snapshot $\mathcal{G}_t$, and $\cdot$ denotes element-wise product.
In addition, to enable a deeper model for learning more implicit information, residual and skip connections are applied throughout the gated HDCC layers.

\subsubsection{Hyperbolic Gated Recurrent Unit}   \label{sec::method:key_modules:hgru}

HGRU~\cite{yangDiscretetimeTemporalNetwork2021} is the hyperbolic variant of GRU~\cite{choLearningPhraseRepresentations2014}.
Performed in the tangent space, HGRU computes the output with high efficiency from historical hidden states and current input features, as 
\begin{equation}
    \textbf{Z}_{i, t} = \exp_{\textbf{o}}^c \left( \text{GRU} \left( \log_{\textbf{o}}^c \left( \textbf{X}_{i, t} \right) , \log_{\textbf{o}}^c \left( \textbf{H}_{i, t - 1} \right) \right) \right).
    \label{eq::method:hgru}
\end{equation}
In HGWaveNet, the historical hidden states are from gated HDCC module, so we use a single HGRU cell for just one step rather than linking it into recursion.

\subsubsection{Hyperbolic Temporal Consistency}   \label{sec::method:key_modules:htc}

In real-world graphs, the evolution proceeds continuously.
Correspondingly, the node representations are expected to change gradually in terms of temporality, which means that the node representations at two consecutive snapshots should keep a short distance.
Hence, HTC~\cite{yangDiscretetimeTemporalNetwork2021} defines a similarity constraint penalty between $\textbf{Z}_{t-1}$ and $\textbf{Z}_t$ at snapshot $\mathcal{G}_t$ as
\begin{equation}
    \mathcal{L}_{\text{HTC}}^t = \frac{1}{N} \sum_{i=1}^N d_c \left( \textbf{Z}_{i, t-1}, \textbf{Z}_{i, t} \right),
    \label{eq::method:htc}
\end{equation}
where $d_c (\cdot)$ is defined in Equation~(\ref{eq::preli:distance}) and $N$ is the number of nodes.
By adding the penalty in optimization process, HTC ensures that the node representations do not change rapidly and the stability of graph evolution is achieved.

\subsubsection{Fermi-Dirac Decoder}   \label{sec::method:key_modules:decoder}

As a generalization of sigmoid, Fermi-Dirac decoder~\cite{krioukovHyperbolicGeometryComplex2010} gives a probability score for edges between nodes $i$ and $j$, which fits the temporal link prediction problem greatly. It is defined as
\begin{equation}
    p_{_{F-D}} (\textbf{x}_i, \textbf{x}_j) = \frac{1}{ \exp \left( \left( d_c \left( \textbf{x}_i, \textbf{x}_j \right) - r \right) / s \right) + 1}, 
    \label{eq::method:decoder}
\end{equation}
where $r$ and $s$ are hyper-parameters, and $\textbf{x}_i, \textbf{x}_j \in \mathbb{B}_c^n$ are hyperbolic representations of nodes $i$ and $j$.

\subsection{Framework of HGWaveNet}   \label{sec::method:framework}

\begin{figure}[t]
    \centering
    \includegraphics[width=0.95\linewidth]{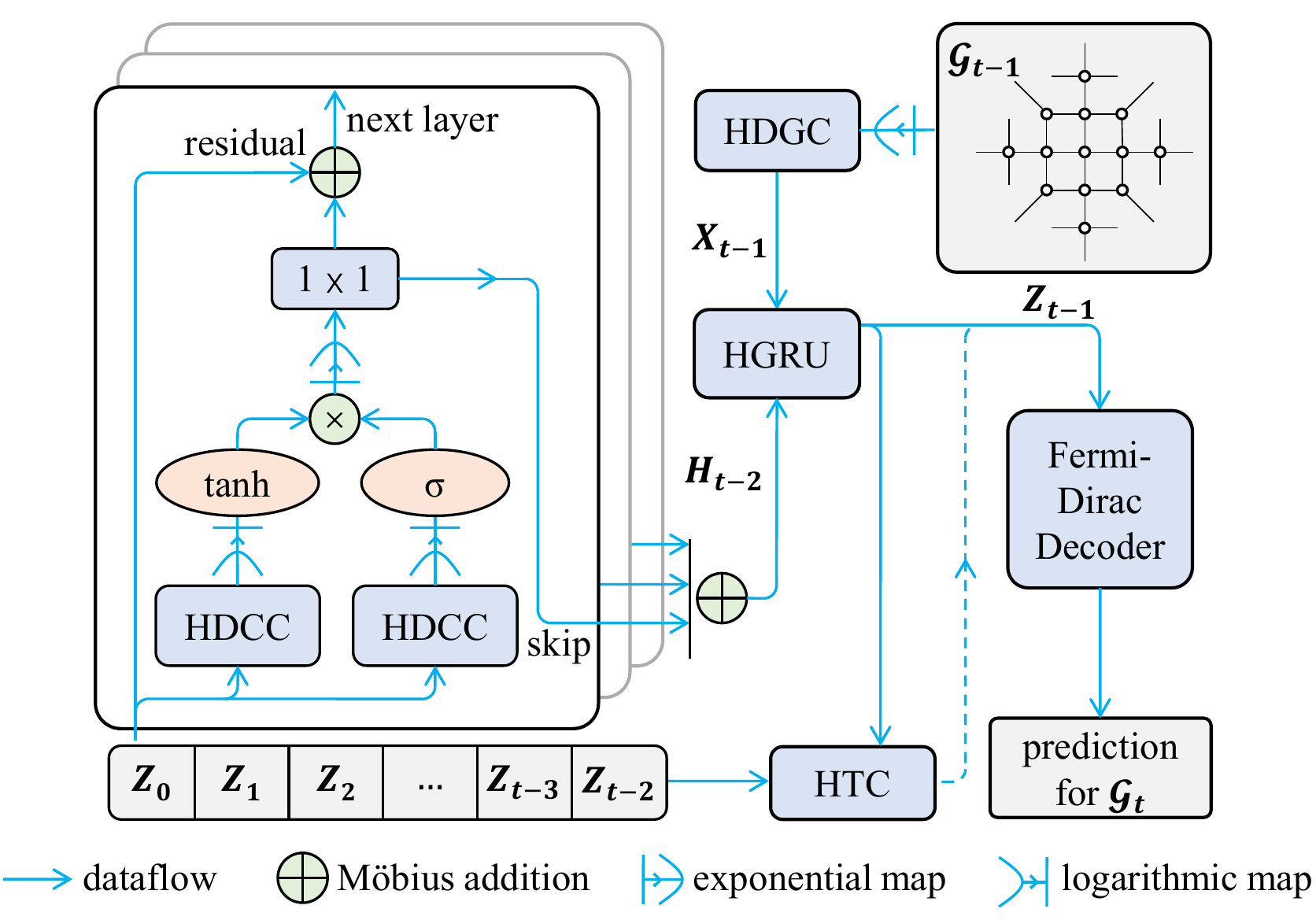}
    \caption{The overall framework of HGWaveNet. For $\mathcal{G}_{t-1}$, the historical evolutionary information handled by gated HDCCs and the spatial topological information handled by HDGC are fed into an HGRU module. With the constraint of HTC, the hyperbolic outputs are decoded by a Fermi-Dirac decoder and the temporal link prediction for $\mathcal{G}_t$ is made.}
    \Description{HGWaveNet.}
    \label{fig::method:hgwavenet}
\end{figure}

With all modules introduced above, we now summarize the overall framework of HGWaveNet in Figure~\ref{fig::method:hgwavenet}.
For the snapshot $\mathcal{G}_{t-1} \in \mathcal{G}$, we first project the input node features into the hyperbolic space with exponential map, and then an HDGC module is applied to learn the spatial topological structure characteristics denoted by $\textbf{X}_{t-1}$.
Simultaneously, the stacked gated HDCC layers calculate the latest historical hidden state $\textbf{H}_{t-2}$ from previous node representations $\textbf{Z}_{t-2}, \textbf{Z}_{t-3}, \cdots$ (the initial historical node representations are padded randomly at the beginning of training).
Next, the spatial information $\textbf{X}_{t-1}$ and the temporal information $\textbf{H}_{t-2}$ are inputted into a single HGRU cell.
The output of HGRU $\textbf{Z}_{t-1}$ is exactly the new latest node representations at snapshot $\mathcal{G}_{t-1}$, and is fed into the Fermi-Dirac decoder to make temporal link predictions for snapshot $\mathcal{G}_t$.

To maximize the probability of connected nodes in $\mathcal{G}_t$ and minimize the probability of unconnected nodes, we use a cross-entropy like loss $\mathcal{L}_{CE}^t$ defined as
\begin{equation}
\begin{split}
    \mathcal{L}_{CE}^t = & \text{Avg}_{i \sim_t j} \left( -\log \left( p_{_{F-D}} \left( \textbf{Z}_{i, t-1}, \textbf{Z}_{j, t-1} \right) \right) \right) + \\
    & \text{Avg}_{i' \nsim_t j'} \left( -\log \left( 1 - p_{_{F-D}} \left( \textbf{Z}_{i', t-1}, \textbf{Z}_{j', t-1} \right) \right) \right),
    \label{eq::method:cross_entropy}
\end{split}
\end{equation}
in which $\sim_t$ denotes the connection at snapshot $\mathcal{G}_t$ between two nodes, and $\nsim_t$ is the opposite.
$(i', j')$ is the sampled negative edge for accelerating training and preventing over-smoothing. 
Taking the HTC module into account and summing up all snapshots, the complete loss function is
\begin{equation}
    \mathcal{L} = \sum_{t=0}^T \left( \mathcal{L}_{CE}^t + \lambda \mathcal{L}_{HTC}^t \right).
    \label{eq::method:loss}
\end{equation}

\paragraph{Time Complexity Analysis}  \label{sec::method:framework:complexity}

We analyse the time complexity of our proposed HGWaveNet by module for each snapshot. 
For the HDGC module, the computation is of $\mathcal{O} ((Nd^2+|\mathcal{E}_t|Kd)L)$, where $d$ is the dimension of node representations, $|\mathcal{E}_t|$ is the edge number for snapshot $\mathcal{G}_t$, $K$ is the truncated diffusion steps and $L$ is the layer number of HDGC.
For gated HDCC, the time complexity is $\mathcal{O} (S^DdD')$, in which $S$ and $D$ are the kernel size and dilated depth of a single HDCC, respectively, and $D'$ is the layer number of the complete gated HDCC module.
HGRU and HTC run only once for each snapshot, and both take the time complexity of $\mathcal{O} (Nd)$.
For Fermi-Dirac decoder, the time complexity is $\mathcal{O} (|\mathcal{E}_t|d+\tilde{|\mathcal{E}_t|}d)$, in which $\tilde{|\mathcal{E}_t|}$ is the number of negative samples for snapshot $\mathcal{G}_t$.

\section{Experiments and Analysis}  \label{sec::exper}

In this section, we conduct extensive experiments on six real-world graphs and prove the superiority of our proposed HGWaveNet. 
In addition, we also conduct ablation studies and hyper-parameter analysis to corroborate the three primary ideas of our method: the fitness between graph distributions and hyperbolic geometry, the effectiveness of spatial information from a wider range of neighbors aggregated by HDGC, and the causality of historical information in the evolutionary process learned by HDCC.

\subsection{Experimental Setup}  \label{sec::exper:setup}

\begin{table}[tbp!]
\centering
\setlength{\abovecaptionskip}{2pt}
\setlength{\belowcaptionskip}{2pt}
\caption{Statistics of datasets.}
\resizebox{0.95\linewidth}{!}{
    \begin{tabular}{c|cccccc}
    \toprule
    \textbf{Datasets} & \textbf{Enron} & \textbf{DBLP} & \textbf{AS733} & \textbf{FB} & \textbf{HepPh} & \textbf{MovieLens} \\
    \midrule
    \textbf{\# Nodes} & 184 & 315 & 6,628 & 45,435 & 15,330 & 9,746 \\
    \textbf{\# Edges} & 790 & 943 & 13,512 & 180,011 & 976,097 & 997,837 \\
    \textbf{\# Snapshots} & 11 & 10 & 30 & 36 & 36 & 11 \\
    \textbf{Train : Test} & 8:3 & 7:3 & 20:10 & 33:3 & 30:6 & 8:3 \\
    \textbf{$\delta$} & 1.5 & 2.0 & 1.5 & 2.0 & 1.0 & 2.0 \\
    \bottomrule
    \end{tabular}
}
\label{tab::exper:datasets}
\end{table}

\subsubsection{Datasets}  \label{sec::exper:setup:datasets}

We evaluate our proposed model and baselines on six real-world datasets from diverse areas, including email communication networks Enron\footnote{\url{https://www.cs.cornell.edu/~arb/data/email-Enron/}}~\cite{bensonSimplicialClosureHigherorder2018}, academic co-author networks DBLP\footnote{\url{https://github.com/VGraphRNN/VGRNN/tree/master/data}}~\cite{hajiramezanaliVariationalGraphRecurrent} and HepPh\footnote{\url{https://snap.stanford.edu/data/cit-HepPh.html}}~\cite{leskovecGraphsTimeDensification2005}, Internet router networks AS733\footnote{\url{https://snap.stanford.edu/data/as-733.html}}~\cite{leskovecGraphsTimeDensification2005}, social networks FB\footnote{\url{https://networkrepository.com/ia-facebook-wall-wosn-dir.php}}~\cite{rossiNetworkDataRepository2015} and movie networks MovieLens\footnote{\url{https://grouplens.org/datasets/movielens/}}~\cite{harperMovieLensDatasetsHistory2016}.
The statistics of these datasets are shown in Table~\ref{tab::exper:datasets}.
We take the same splitting ratios for training and testing as ~\cite{yangDiscretetimeTemporalNetwork2021} on all datasets except MovieLens on which a similar manner is adopted.
Gromov's hyperbolicity~\cite{gromovHyperbolicGroups} $\delta$ is used to measure the tree-likeness and hierarchical properties of graphs. 
A lower $\delta$ denotes a more tree-like structure and $\delta=0$ denotes a pure tree.
The datasets we choose all remain implicitly hierarchical and show distinct power-law distributions.

\begin{table}[tbp!]
\centering
\setlength{\tabcolsep}{6pt}
\setlength{\abovecaptionskip}{2pt}
\setlength{\belowcaptionskip}{2pt}
\caption{Baselines.}
\resizebox{0.67\linewidth}{!}{
    \begin{tabular}{c|cc}
    \toprule
    \textbf{Methods} & \textbf{Static/dynamic} & \textbf{Manifolds} \\
    \midrule
    \textbf{HGCN}~\cite{chamiHyperbolicGraphConvolutional} & Static & Lorentz \\
    \textbf{HAT}\cite{zhangHyperbolicGraphAttention2021} & Static & Poincaré \\
    \textbf{EvolveGCN}~\cite{parejaEvolveGCNEvolvingGraph2020} & Dynamic & Euclidean \\
    \textbf{GRUGCN}~\cite{seoStructuredSequenceModeling2016} & Dynamic & Euclidean \\
    \textbf{TGN}~\cite{rossiTemporalGraphNetworks2020a} & Dynamic & Euclidean \\
    \textbf{DySAT}~\cite{sankarDySATDeepNeural2020} & Dynamic & Euclidean \\
    \textbf{HTGN}~\cite{yangDiscretetimeTemporalNetwork2021} & Dynamic & Poincaré \\
    \bottomrule
    \end{tabular}
}
\label{tab::exper:baselines}
\end{table}

\subsubsection{Baselines}  \label{sec::exper:setup:baselines}

Considering that HGWaveNet is constructed in the hyperbolic space for dynamic graphs, we choose seven competing baselines either in hyperbolic spaces or built for dynamic graphs to verify the superiority of our model.
The baselines are summarized in Table~\ref{tab::exper:baselines}, where HTGN on Poincaré ball shows state-of-the-art performance in most evaluations.

\begin{table*}[tbp!]
\centering
\setlength{\tabcolsep}{3pt}
\setlength{\abovecaptionskip}{2pt}
\setlength{\belowcaptionskip}{2pt}
\caption{Performance on temporal link prediction.}
\resizebox{0.98\linewidth}{!}{
    \begin{tabular}{cc|cc|cc|cc|cc|cc|cc}
    \toprule
    & \textbf{Dataset} & \multicolumn{2}{c|}{\textbf{Enron}} & \multicolumn{2}{c|}{\textbf{DBLP}} & \multicolumn{2}{c|}{\textbf{AS733}} & \multicolumn{2}{c|}{\textbf{FB}} & \multicolumn{2}{c|}{\textbf{HepPh}} & \multicolumn{2}{c}{\textbf{MovieLens}} \\
    & \textbf{Metric} & \textbf{AUC} & \textbf{AP} & \textbf{AUC} & \textbf{AP} & \textbf{AUC} & \textbf{AP} & \textbf{AUC} & \textbf{AP} & \textbf{AUC} & \textbf{AP} & \textbf{AUC} & \textbf{AP} \\
    \midrule
    \multirow{7}{*}{\rotatebox{90}{\textbf{Baselines}}} & \textbf{HGCN} & $93.96 \pm 0.09$ & $93.55 \pm 0.06$ & $89.16 \pm 0.16$ & $91.63 \pm 0.22$ & $96.34 \pm 0.05$ & $93.28 \pm 0.11$ & \underline{$86.11 \pm 0.13$} & $83.74 \pm 0.15$ & $90.64 \pm 0.07$ & $88.98 \pm 0.09$ & $84.89 \pm 0.40$ & $77.98 \pm 0.66$ \\
    & \textbf{HAT} & \underline{$94.49 \pm 0.11$} & \underline{$94.63 \pm 0.08$} & \underline{$89.29 \pm 0.18$} & $90.15 \pm 0.14$ & $96.09 \pm 0.01$ & $93.21 \pm 0.02$ & $84.02 \pm 0.09$ & $83.03 \pm 0.15$ & $90.52 \pm 0.04$ & \underline{$89.53 \pm 0.04$} & $86.26 \pm 0.14$ & $84.19 \pm 0.21$ \\
    & \textbf{EvolveGCN} & $90.12 \pm 0.69$ & $92.71 \pm 0.34$ & $83.88 \pm 0.53$ & $87.53 \pm 0.22$ & $92.47 \pm 0.04$ & $95.28 \pm 0.01$ & $76.85 \pm 0.85$ & $80.87 \pm 0.64$ & $76.82 \pm 1.46$ & $81.18 \pm 0.89$ & $55.41 \pm 1.55$ & $69.41 \pm 1.00$ \\
    & \textbf{GRUGCN} & $92.47 \pm 0.36$ & $93.38 \pm 0.24$ & $84.60 \pm 0.92$ & $87.87 \pm 0.58$ & $94.96 \pm 0.35$ & $96.64 \pm 0.22$ & $79.38 \pm 1.02$ & $82.77 \pm 0.75$ & $82.86 \pm 0.53$ & $85.87 \pm 0.23$ & $71.01 \pm 0.77$ & $79.04 \pm 0.53$ \\
    & \textbf{TGN} & $91.77 \pm 1.04$ & $85.36 \pm 0.87$ & $87.20 \pm 0.82$ & $84.87 \pm 0.80$ & $93.19 \pm 1.03$ & $92.27 \pm 0.83$ & $80.85 \pm 0.23$ & $81.92 \pm 1.34$ & $74.78 \pm 0.47$ & $74.61 \pm 0.43$ & $71.52 \pm 0.59$ & $70.53 \pm 0.38$ \\
    & \textbf{DySAT} & $93.06 \pm 0.97$ & $93.06 \pm 1.05$ & $87.25 \pm 1.70$ & $90.40 \pm 1.47$ & $95.06 \pm 0.21$ & $96.72 \pm 0.12$ & $76.88 \pm 0.08$ & $80.39 \pm 0.14$ & $81.02 \pm 0.25$ & $84.47 \pm 0.23$ & $70.21 \pm 0.58$ & $78.41 \pm 0.33$ \\
    & \textbf{HTGN} & $94.17 \pm 0.17$ & $94.31 \pm 0.26$ & $89.26 \pm 0.17$ & \underline{$91.91 \pm 0.07$} & \underline{$98.75 \pm 0.03$} & \underline{$98.41 \pm 0.03$} & $83.70 \pm 0.33$ & \underline{$83.80 \pm 0.43$} & \underline{$91.13 \pm 0.14$} & $89.52 \pm 0.28$ & \underline{$86.80 \pm 0.34$} & \underline{$85.40 \pm 0.19$} \\
    \midrule
    \multirow{2}{*}{\rotatebox{90}{\textbf{Ours}}} & \textbf{HGWaveNet} & \bm{$96.86 \pm 0.08$} & \bm{$97.04 \pm 0.07$} & \bm{$89.96 \pm 0.27$} & \bm{$92.12 \pm 0.18$} & \bm{$98.78 \pm 0.01$} & \bm{$98.53 \pm 0.02$} & \bm{$89.51 \pm 0.28$} & \bm{$86.88 \pm 0.29$} & \bm{$92.37 \pm 0.04$} & \bm{$91.48 \pm 0.05$} & \bm{$91.90 \pm 0.06$} & \bm{$90.65 \pm 0.12$} \\
    & \textbf{Gain(\%)} & \textbf{+2.51} & \textbf{+2.55} & \textbf{+0.75} & \textbf{+0.23} & \textbf{+0.03} & \textbf{+0.12} & \textbf{+3.95} & \textbf{+3.68} & \textbf{+1.36} & \textbf{+2.18} & \textbf{+5.88} & \textbf{+6.15} \\
    \midrule
    \multirow{6}{*}{\rotatebox{90}{\textbf{Ablation}}} & \textbf{w/o HDGC} & $94.95 \pm 0.10$ & $95.24 \pm 0.07$ & $89.51 \pm 0.24$ & $91.84 \pm 0.17$ & $97.97 \pm 0.20$ & $97.35 \pm 0.24$ & $85.99 \pm 0.29$ & $82.77 \pm 0.48$ & $90.61 \pm 0.22$ & $89.68 \pm 0.03$ & $84.35 \pm 0.12$ & $82.77 \pm 0.38$ \\
    & \textbf{Gain(\%)} & -1.97 & -1.85 & -0.50 & -0.30 & -0.82 & -1.20 & -3.93 & -4.73 & -1.91 & -1.97 & -8.22 & -8.69 \\
    \cmidrule(lr){2-14}
    & \textbf{w/o HDCC} & $95.33 \pm 0.09$ & $95.64 \pm 0.09$ & $89.56 \pm 0.22$ & $92.04 \pm 0.12$ & $97.83 \pm 0.17$ & $97.25 \pm 0.21$ & $85.13 \pm 0.16$ & $82.38 \pm 0.88$ & $90.83 \pm 0.20$ & $89.77 \pm 0.32$ & $82.46 \pm 0.37$ & $80.29 \pm 0.58$ \\
    & \textbf{Gain(\%)} & -1.58 & -1.44 & -0.44 & -0.09 & -0.96 & -1.30 & -4.89 & -5.18 & -1.67 & -1.87 & -10.27 & -11.43 \\
    \cmidrule(lr){2-14}
    & \textbf{w/o $\mathbb{B}$} & $92.62 \pm 0.55$ & $93.77 \pm 0.33$ & $85.18 \pm 0.29$ & $87.76 \pm 0.11$ & $95.21 \pm 0.11$ & $96.85 \pm 0.08$ & $79.20 \pm 0.34$ & $82.45 \pm 0.21$ & $76.31 \pm 0.75$ & $78.11 \pm 0.29$ & $73.70 \pm 0.54$ & $80.15 \pm 0.54$ \\
    & \textbf{Gain(\%)} & -4.38 & -3.37 & -5.31 & -4.73 & -3.61 & -1.71 & -11.52 & -5.10 & -17.39 & -14.62 & -19.80 & -11.58 \\
    \bottomrule
    \end{tabular}
}
\footnotesize
\flushleft{{$^*$ For evaluations on baselines and our model, the best results are in bold, and the suboptimal results are underlined. The same applies to Table~\ref{tab::exper:new_link}.}}
\flushleft{{$^{\star}$ Part of results are from~\cite{yangDiscretetimeTemporalNetwork2021} under the same experimental setup. The same applies to Table~\ref{tab::exper:new_link}.}}
\label{tab::exper:link}
\end{table*}

\begin{table*}[tbp!]
\centering
\setlength{\tabcolsep}{3pt}
\setlength{\abovecaptionskip}{2pt}
\setlength{\belowcaptionskip}{2pt}
\caption{Performance on temporal new link prediction.}
\resizebox{0.98\linewidth}{!}{
    \begin{tabular}{cc|cc|cc|cc|cc|cc|cc}
    \toprule
    & \textbf{Dataset} & \multicolumn{2}{c|}{\textbf{Enron}} & \multicolumn{2}{c|}{\textbf{DBLP}} & \multicolumn{2}{c|}{\textbf{AS733}} & \multicolumn{2}{c|}{\textbf{FB}} & \multicolumn{2}{c|}{\textbf{HepPh}} & \multicolumn{2}{c}{\textbf{MovieLens}} \\
    & \textbf{Metric} & \textbf{AUC} & \textbf{AP} & \textbf{AUC} & \textbf{AP} & \textbf{AUC} & \textbf{AP} & \textbf{AUC} & \textbf{AP} & \textbf{AUC} & \textbf{AP} & \textbf{AUC} & \textbf{AP} \\
    \midrule
    \multirow{7}{*}{\rotatebox{90}{\textbf{Baselines}}} & \textbf{HGCN} & $90.36 \pm 0.16$ & $88.33 \pm 0.55$ & $81.20 \pm 0.19$ & $83.28 \pm 0.23$ & $92.33 \pm 0.12$ & $88.31 \pm 0.16$ & $81.04 \pm 0.14$ & $80.59 \pm 0.13$ & $89.64 \pm 0.27$ & $87.87 \pm 0.11$ & $85.27 \pm 0.35$ & $78.56 \pm 0.49$ \\
    & \textbf{HAT} & $90.34 \pm 0.10$ & $88.60 \pm 0.19$ & $79.29 \pm 0.15$ & $82.58 \pm 0.08$ & $91.65 \pm 0.08$ & $90.13 \pm 0.15$ & \underline{$83.05 \pm 0.10$} & \underline{$82.96 \pm 0.18$} & $89.63 \pm 0.05$ & \underline{$88.34 \pm 0.04$} & $86.54 \pm 0.16$ & $84.60 \pm 0.25$ \\
    & \textbf{EvolveGCN} & $82.85 \pm 0.97$ & $85.01 \pm 0.22$ & $73.49 \pm 0.86$ &  $77.11 \pm 0.44$ & $75.82 \pm 0.67$ & $83.57 \pm 0.46$ & $74.49 \pm 0.89$ & $78.33 \pm 0.66$ & $74.79 \pm 1.61$ & $79.04 \pm 1.02$ & $55.45 \pm 1.53$ & $69.35 \pm 0.98$ \\
    & \textbf{GRUGCN} & $87.59 \pm 0.57$ & $88.41 \pm 0.45$ & $75.60 \pm 1.60$ & $78.55 \pm 1.05$ & $83.14 \pm 1.21$ & $88.14 \pm 0.76$ & $77.69 \pm 1.03$ & $81.07 \pm 0.77$ & $81.97 \pm 0.49$ & $84.78 \pm 0.22$ & $71.32 \pm 0.77$ & $79.51 \pm 0.50$ \\
    & \textbf{TGN} & $86.75 \pm 0.59$ & $81.37 \pm 0.42$ & \underline{$81.83 \pm 1.08$} & $81.61 \pm 0.62$ & $87.89 \pm 0.63$ & $84.94 \pm 0.94$ & $78.93 \pm 0.98$ & $76.57 \pm 0.63$ & $74.21 \pm 0.78$ & $76.12 \pm 0.54$ & $72.22 \pm 0.68$ & $70.90 \pm 0.40$ \\
    & \textbf{DySAT} & $87.94 \pm 3.78$ & $86.83 \pm 5.01$ & $79.74 \pm 4.35$ & $83.47 \pm 3.01$ & $82.84 \pm 0.72$ & $89.07 \pm 0.57$ & $74.97 \pm 0.12$ & $78.34 \pm 0.07$ & $79.01 \pm 0.26$ & $82.53 \pm 0.25$ & $70.63 \pm 0.58$ & $79.02 \pm 0.33$ \\
    & \textbf{HTGN} & \underline{$91.26 \pm 0.27$} & \underline{$90.62 \pm 0.34$} & $81.74 \pm 0.56$ & \underline{$84.06 \pm 0.41$} & \underline{$96.62 \pm 0.22$} & \underline{$95.52 \pm 0.25$} & $82.21 \pm 0.41$ & $81.70 \pm 0.46$ & \underline{$90.11 \pm 0.14$} & $88.18 \pm 0.31$ & \underline{$87.06 \pm 0.25$} & \underline{$85.82 \pm 0.29$} \\
    \midrule
    \multirow{2}{*}{\rotatebox{90}{\textbf{Ours}}} & \textbf{HGWaveNet} & \bm{$93.49 \pm 0.08$} & \bm{$92.51 \pm 0.10$} & \bm{$84.26 \pm 0.46$} & \bm{$86.19 \pm 0.33$} & \bm{$96.99 \pm 0.07$} & \bm{$95.60 \pm 0.10$} & \bm{$88.59 \pm 0.28$} & \bm{$86.00 \pm 0.30$} & \bm{$91.45 \pm 0.05$} & \bm{$90.21 \pm 0.04$} & \bm{$92.04 \pm 0.04$} & \bm{$90.66 \pm 0.07$} \\
    & \textbf{Gain(\%)} & \textbf{+2.44} & \textbf{+2.09} & \textbf{+2.97} & \textbf{+2.53} & \textbf{+0.38} & \textbf{+0.08} & \textbf{+6.67} & \textbf{+3.66} & \textbf{+1.49} & \textbf{+2.12} & \textbf{+5.72} & \textbf{+5.64} \\
    \midrule
    \multirow{6}{*}{\rotatebox{90}{\textbf{Ablation}}} & \textbf{w/o HDGC} & $90.67 \pm 0.20$ & $89.90 \pm 0.22$ & $83.47 \pm 0.36$ & $85.85 \pm 0.28$ & $95.69 \pm 0.34$ & $93.99 \pm 0.36$ & $84.96 \pm 0.30$ & $81.57 \pm 0.42$ & $89.63 \pm 0.26$ & $88.27 \pm 0.03$ & $84.76 \pm 0.15$ & $83.42 \pm 0.33$ \\
    & \textbf{Gain(\%)} & -3.02 & -2.82 & -0.94 & -0.39 & -1.34 & -1.68 & -4.10 & -5.15 & -1.99 & -2.15 & -7.91 & -7.99 \\
    \cmidrule(lr){2-14}
    & \textbf{w/o HDCC} & $91.76 \pm 0.08$ & $90.87 \pm 0.06$ & $82.36 \pm 0.23$ & $84.87 \pm 0.10$ & $94.56 \pm 0.44$ & $92.91 \pm 0.45$ & $84.03 \pm 0.14$ & $80.99 \pm 0.76$ & $89.87 \pm 0.20$ & $88.42 \pm 0.30$ & $82.92 \pm 0.31$ & $81.23 \pm 0.50$ \\
    & \textbf{Gain(\%)} & -1.85 & -1.77 & -2.25 & -1.53 & -2.51 & -2.81 & -5.15 & -5.83 & -1.73 & -1.98 & -9.91 & -10.40 \\
    \cmidrule(lr){2-14}
    & \textbf{w/o $\mathbb{B}$} & $87.10 \pm 0.72$ & $88.49 \pm 0.57$ & $76.65 \pm 0.80$ & $79.53 \pm 0.67$ & $84.85 \pm 0.77$ & $89.59 \pm 0.37$ & $77.37 \pm 0.33$ & $80.63 \pm 0.19$ & $74.23 \pm 0.58$ & $76.21 \pm 0.53$ & $74.11 \pm 0.55$ & $80.51 \pm 0.31$ \\
    & \textbf{Gain(\%)} & -6.83 & -4.35 & -9.03 & -7.73 & -12.52 & -6.29 & -12.67 & -6.24 & -18.83 & -15.52 & -19.48 & -11.20 \\
    \bottomrule
    \end{tabular}
}
\footnotesize
\label{tab::exper:new_link}
\end{table*}

\subsubsection{Evaluation Tasks and Metrics}  \label{sec::exper:setup:tasks}

Similar to ~\cite{yangDiscretetimeTemporalNetwork2021}, our experiments consist of two different tasks: \textit{temporal link prediction} and \textit{temporal new link prediction}.
Specifically, \textit{temporal link prediction} aims to predict the edges that appear in $\mathcal{G}_t$ and other snapshots after timestamp $t$ based on the training on $(\mathcal{G}_0, \mathcal{G}_1, \cdots, \mathcal{G}_{t-1})$, while \textit{temporal new link prediction} aims to predict those in $\mathcal{G}_t$ but not in $\mathcal{G}_{t-1}$. 
To quantify the experimental performance, we choose the widely used metrics average precision (AP) and area under ROC curve (AUC). 
The datasets are split into training sets and test sets as shown in Table~\ref{tab::exper:datasets} by snapshots to run both our proposed model and baselines on the above tasks.



\subsection{Experimental Results}  \label{sec::expr:results}

We implement HGWaveNet\footnote{The code is open sourced in \url{https://github.com/TaiLvYuanLiang/HGWaveNet}.} with PyTorch on Ubuntu 18.04.
Each experiment is repeated five times to avoid random errors, and the average results (with form \textit{average value $\pm$ standard deviation}) on the test sets are reported in Table~\ref{tab::exper:link} and Table~\ref{tab::exper:new_link}.
Next, we discuss the experimental results on each evaluation task separately.

\subsubsection{Temporal Link Prediction}  \label{sec::expr:results:temporal}

The results on temporal link prediction are shown in Table~\ref{tab::exper:link}. 
Obviously, our HGWaveNet outperforms all baselines including the hyperbolic state-of-the-art model HTGN both on AUC and AP. 
To analyse the results further, we intuitively divide the six datasets into two groups: small (Enron, DBLP, AS733) and large (FB, HepPh, MovieLens), according to the scales.
It can be observed that the superiority of our model is more significant on large graphs than small ones compared with baselines, especially those Euclidean methods. 
This is because the representation capacity of Euclidean methods decreases rapidly with increasing graph scales, while hyperbolic models remain stable by virtue of the fitness between data distributions and space properties.

\subsubsection{Temporal New Link Prediction}  \label{sec::expr:results:new}

The results on temporal new link prediction are shown in Table~\ref{tab::exper:new_link}. 
This task aims to predict the edges unseen in the training process and evaluates the inductive ability of models.
Our model shows greater advantages, and similar gaps with those on temporal link prediction between small and large graphs appear again.
Another observed fact is that even for the inductive evaluation task, the two hyperbolic static methods HGCN and HAT have a relatively better performance than the Euclidean dynamic models.
This observation further demonstrates the excellence of modeling real-world graphs on hyperbolic spaces.

\subsection{Ablation Study}  \label{sec::expr:ablation}

To asses the contribution of each component, we conduct the following ablation study on three HGWaveNet variants:
\begin{itemize}
\item \textbf{w/o HDGC:} HGWaveNet without the hyperbolic diffusion graph convolution, implemented by replacing HDGC with shallow HGCN to learn from spatial characteristics.
\item \textbf{w/o HDCC:} HGWaveNet without the hyperbolic dilated causal convolution, implemented by replacing HDCC with an attention mechanism to achieve historical hidden states.
\item \textbf{w/o $\mathbb{B}$:} HGWaveNet without hyperbolic geometry, implemented by replacing all modules with corresponding Euclidean versions.
\end{itemize}

We take the same experimental setup with HGWaveNet and baselines on these three variants, and the results are reported in the last rows of Table~\ref{tab::exper:link} and Table~\ref{tab::exper:new_link}.
The performance drops noticeably after removing any of these three components, which indicates their respective importance.
In the following, we discuss the details of the effectiveness of these variants.


HDGC module imposes the diffusion process into hyperbolic graph convolution networks.
It allows nodes to aggregate information from a wider range of neighbors and improves the efficiency of message propagation by calculating the stationary distribution of a long Markov process in a closed form.
The fact that performance of w/o HDGC variant has a larger degradation on large graphs than small ones forcefully proves the effectiveness of this component for large graphs with long paths.


The purpose of HDCC module is to capture the internal order of causal correlation in sequential data. 
Generally, the larger a graph is, the more complex its evolution is and the richer historical information it has.
The experimental results on w/o HDCC show the powerful ability of this component to capture causal order in temporal evolution process, especially on complex graphs (i.e. FB and MovieLens).


As observed in Table~\ref{tab::exper:link} and Table~\ref{tab::exper:new_link}, the degradation of w/o $\mathbb{B}$ is much more severe than that of the other two variants, which strongly supports our primary idea that hyperbolic geometry fits the power-law distributions of real-world graphs very well.
Moreover, it is noteworthy that for most results, the variant w/o $\mathbb{B}$ exhibits a comparable or even exceeding performance to other Euclidean baselines.
This proves that our model architecture still keeps advanced even without the advantages of hyperbolic geometry.

\subsection{Parameter Analysis}  \label{sec::expr:para}

We further study the influences of three hyper-parameters: representation dimension, truncated diffusion step and dilated depth.

\begin{figure}[t]
    \centering
    \includegraphics[width=0.95\linewidth]{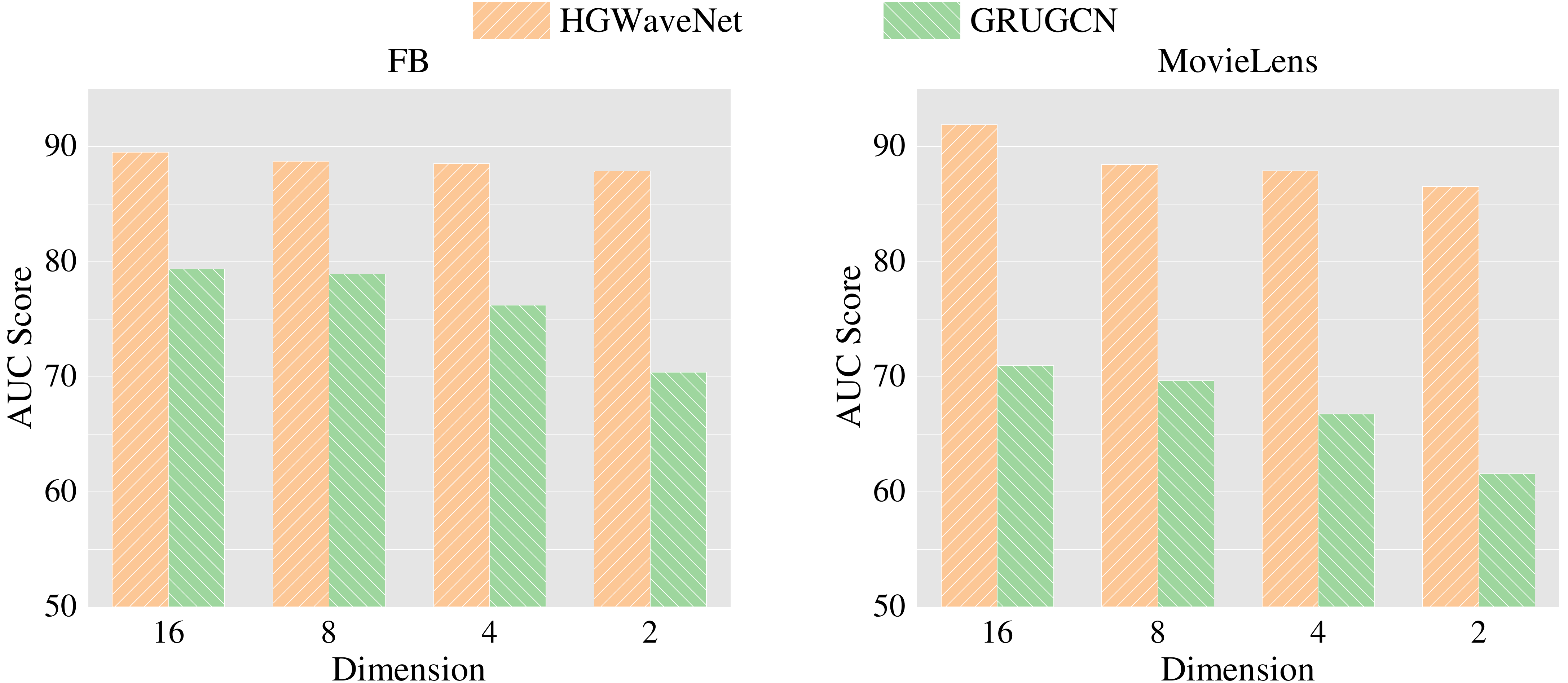}
    \caption{The influence of representation dimension on two datasets FB and MovieLens for HGWaveNet and GRUGCN.}
    \Description{Influence of dimension.}
    \label{fig::exper:dim_parameter}
\end{figure}

We evaluate the performance of HGWaveNet and one of the best Euclidean models GRUGCN on two datasets FB (relatively sparse) and MovieLens (relatively dense) by setting the representation dimension into different values, as shown in Figure~\ref{fig::exper:dim_parameter}.
On the one hand, it is clear that with the dimension decreasing, our model remains stable compared to GRUGCN, to which the hyperbolic geometry contributes greatly.
On the other hand, for HGWaveNet, the degradation of the performance on MovieLens is severer than that on FB, which is consistent with the conclusion in ~\cite{zhangWhereAreWe2021} that hyperbolic models fit sparse data better than dense data.

\begin{figure}[t]
    \centering
    \includegraphics[width=0.42\linewidth]{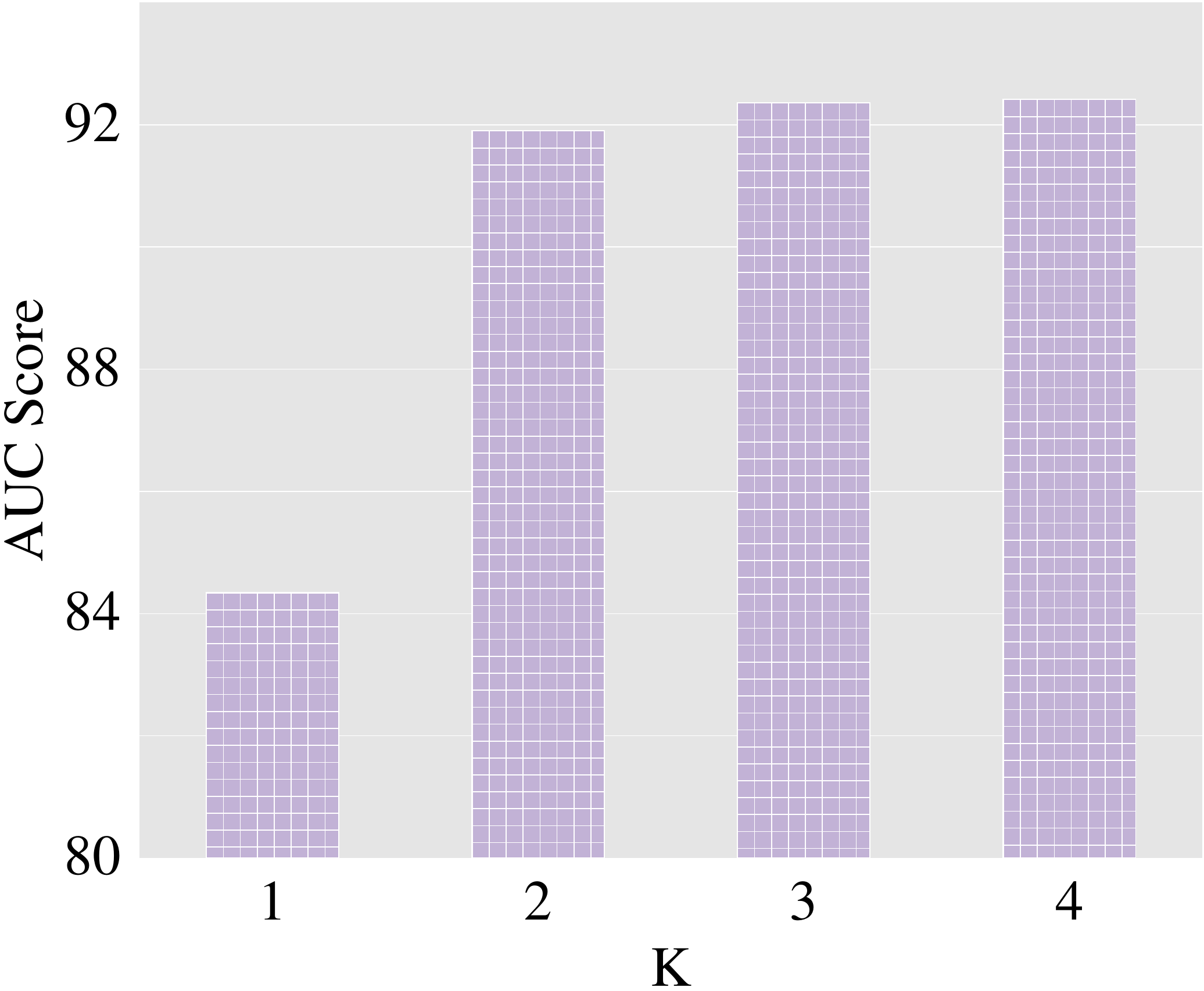}
    \hspace{0.05\linewidth}
    \includegraphics[width=0.42\linewidth]{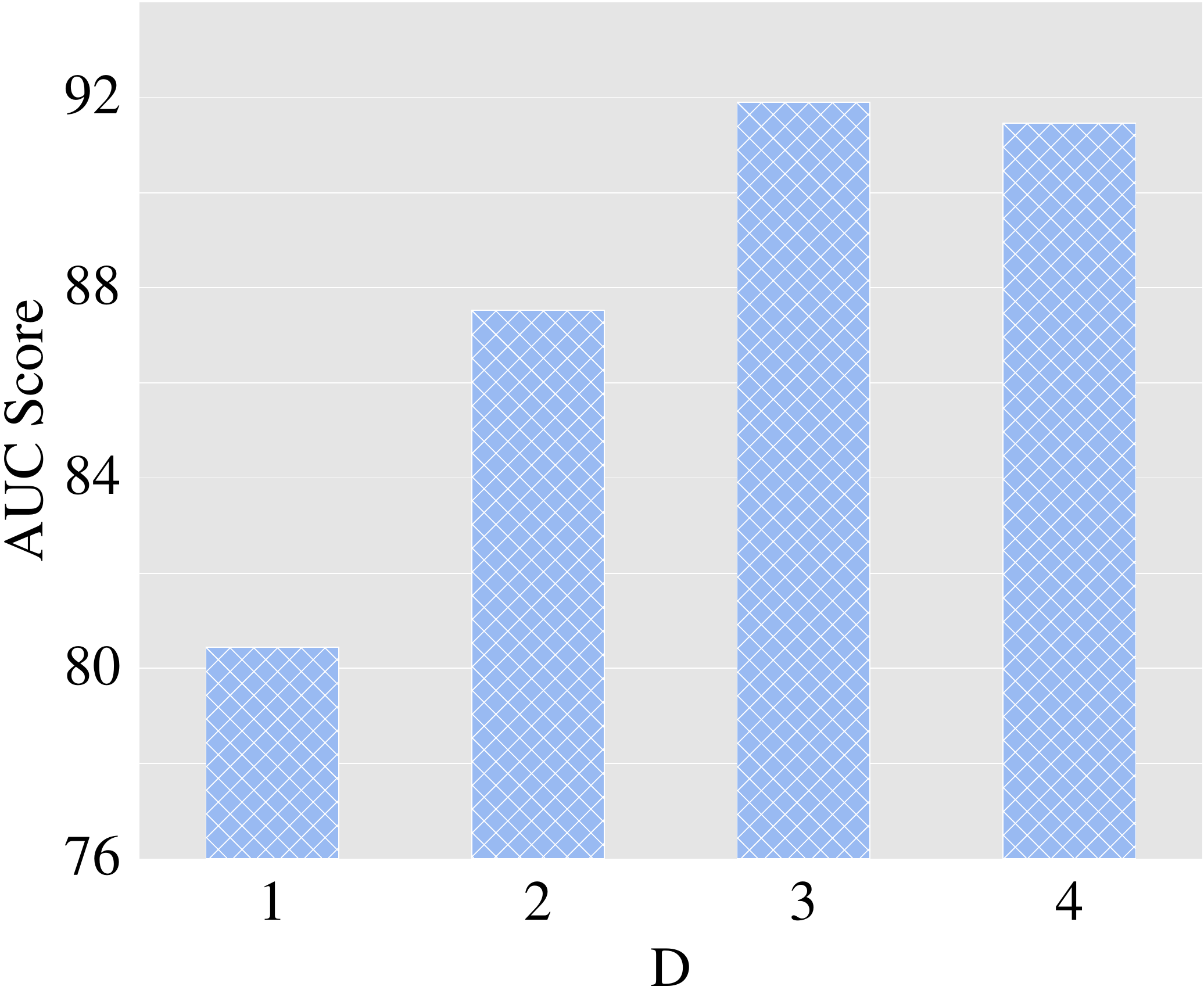}
    \caption{The influence of truncated diffusion step $K$ (left) and dilated depth $D$ (right) on MovieLens.}
    \Description{Influence of truncated diffusion step and dilated depth.}
    \label{fig::exper:D_K_parameter}
\end{figure}

Figure~\ref{fig::exper:D_K_parameter} shows the results of our model for temporal link prediction on MovieLens with different values of truncated diffusion step $K$ in HDGC and dilated depth $D$ in HDCC.
The wrap-up observation that the performance generally increases as $K$ and $D$ increase proves the positive influences of diffusion process and dilation convolution.
Specifically, in terms of truncated diffusion step, the performance has a great improvement from $K=1$ to $K=2$.
When $K > 2$, the benefit from increasing $K$ becomes too small to match the surge of computation from exponentially expanding neighbors in graphs.
Dilated depth $D$ is used to control the receptive field of causal convolution.
However, as the receptive field expands, more noise is imported and may degrade the model performance.
According to our extensive experiments, $D=3$ is a nice choice in most situations, with the causal convolution kernel size being 2. 

\section{Conclusion}  \label{sec::conclusion}

In this paper, we propose a hyperbolic graph neural network HGWaveNet for temporal link prediction.
Inspired by the observed power-law distribution and implicit hierarchical structure of real-world graphs, we construct our model on the Poincaré ball, one of the isometric models of hyperbolic spaces.
Specifically, we design two novel modules hyperbolic diffusion graph convolution (HDGC) and hyperbolic dilated causal convolution (HDCC) to extract the spatial topological information and temporal evolutionary states, respectively.
HDGC imposes the diffusion process into graph convolution and provides an efficient way to aggregate information from a wider range of neighbors.
HDCC ensures that the internal order of causal correlations is not violated by applying the convolution filter only to previous data.
Extensive experiments on diverse real-world datasets prove the superiority of HGWaveNet, and the ablation study further verifies the effectiveness of each component of our model.
In future work, we will further generalize our model for more downstream tasks and try to use hyperbolic GNNs to capture more complex semantic information in heterogeneous graphs.

\begin{acks}
This work is supported by the Chinese Scientific and Technical Innovation Project 2030 (2018AAA0102100), National Natural Science Foundation of China (U1936206, 62172237) and the Tianjin Natural Science Foundation for Distinguished Young Scholars (22JCJQJC00150).
\end{acks}


\bibliographystyle{ACM-Reference-Format}
\bibliography{biblist}

\appendix
\section{Appendix} \label{sec::appendix}

\subsection{Notations}  \label{sec::appendix:notations}
Part of notations in our paper are summarized in Table~\ref{tab::appendix:notations}.

\begin{table}[H]
\centering
\setlength{\tabcolsep}{6pt}
\setlength{\abovecaptionskip}{2pt}
\setlength{\belowcaptionskip}{2pt}
\caption{Notations.}
\resizebox{0.9\linewidth}{!}{
    \begin{tabular}{c|l}
    \toprule
    \textbf{Symbols} & \multicolumn{1}{c}{\textbf{Descriptions}} \\
    \midrule
    $\mathcal{G}$ & Discrete dynamic graph represented by snapshots. \\
    $\mathcal{G}_t$ & Snapshot of $\mathcal{G}$ at timestamp $t$. \\
    $\mathbb{H}_c^n$ & $n$-dimensional hyperbolic manifold with curvature $-c$. \\
    $g^{c, \mathbb{H}}$ & Riemannian metric corresponding to the manifold $\mathbb{H}$. \\
    $\mathcal{T}_{\mathbf{x}}\mathbb{H}_c^n$ & Tangent space of $\mathbb{H}_c^n$ at point $\mathbf{x}$. \\
    $\mathbb{B}_c^n$ & Poincaré ball manifold corresponding to $\mathbb{H}_c^n$. \\
    $d$ & Dimension of node representations. \\
    $K$ & Truncated diffusion steps in HDGC. \\
    $L$ & Layer number of HDGC. \\
    $S$ &  Convolution kernel size of HDCC. \\
    $D$ & Dilated depth of HDCC. \\
    $D'$ & Layer number of gated HDCC module. \\
    $\mathbf{A}$ & Bi-direct adjacent matrix of a static graph or snapshot. \\
    $\mathbf{X}_{t, k}^l$ & $k$-th step of HDGC at layer $l$, timestamp $t$. \\
    $\mathbf{H}_t$ & Temporal information earlier than timestamp $t$. \\
    $\mathbf{F}$ & Kernel metrix of HDCC. \\ 
    $\mathbf{Z}_t$ & Node representations at timestamp $t$. \\
    $\mathcal{L}_{\text{HTC}}^t$ & Hyperbolic temporal consistency at timestamp $t$. \\
    $\mathcal{L}_{\text{CE}}^t$ &  Cross-entropy like loss at timestamp $t$. \\
    $\mathbf{W}, \mathbf{b}$ & Trainable parameters. \\
    \bottomrule
    \end{tabular}
}
\label{tab::appendix:notations}
\end{table}

\subsection{Experimental Details}  \label{sec::appendix:details}

For all baselines, we take the recommended parameter settings unless otherwise noted.
For our proposed HGWaveNet, the parameters are set as follows: the truncated diffusion step $K=2$, the layer number of HDGC $L=2$, the dilation depth for HDCC $D=3$, kernel size $S=2$, the layer number for gated HDCC is valued as 4, and the hyper-parameters $r, s$ in Fermi-Dirac decoder are taken to be 2 and 1 separately.
All involved curvatures are initialized as 1 and keep trainable during the training process.
For fairness, the representation dimension is set to 16 for all methods.
All experiments are run on a machine with 2 $\times$ Intel Xeon Gold 6226R 16C 2.90 GHz CPUs, 4 $\times$ GeForce RTX 3090 GPUs.

\subsection{Supplementary Experiments}  \label{sec::appendix:para}

In Table~\ref{tab::appendix:dim_para}, ~\ref{tab::appendix:step_para} and ~\ref{tab::appendix:depth_para}, we give out the supplementary experimental results for parameter analysis in Section~\ref{sec::expr:para}.  All these experiments are evaluated by AUC score on temporal link prediction task.

\begin{table}[H]
\centering
\setlength{\tabcolsep}{6pt}
\setlength{\abovecaptionskip}{2pt}
\setlength{\belowcaptionskip}{2pt}
\caption{The influence of representation dimension.}
\resizebox{0.9\linewidth}{!}{
    \begin{tabular}{c|cccccc}
    \toprule
    \textbf{Dimension} & \textbf{Enron} & \textbf{DBLP} & \textbf{AS733} & \textbf{FB} & \textbf{HepPh} & \textbf{MovieLens} \\
    \midrule
    \textbf{16} & 96.86 & 89.96 & 98.78 & 89.51 & 92.37 & 91.90 \\
    \textbf{8} & 96.56 & 89.49 & 98.67 & 88.74 & 90.20 & 88.44 \\
    \textbf{4} & 96.64 & 89.24 & 98.36 & 88.51 & 88.76 & 87.90 \\
    \textbf{2} & 96.41 & 88.84 & 97.95 & 87.87 & 88.42 & 86.55 \\
    \bottomrule
    \end{tabular}
}
\label{tab::appendix:dim_para}
\end{table}

\begin{table}[H]
\centering
\setlength{\tabcolsep}{6pt}
\setlength{\abovecaptionskip}{2pt}
\setlength{\belowcaptionskip}{2pt}
\caption{The influence of truncated diffusion step $K$.}
\resizebox{0.9\linewidth}{!}{
    \begin{tabular}{c|cccccc}
    \toprule
    $K$ & \textbf{Enron} & \textbf{DBLP} & \textbf{AS733} & \textbf{FB} & \textbf{HepPh} & \textbf{MovieLens} \\
    \midrule
    \textbf{1} & 94.95 & 89.51 & 97.97 & 85.99 & 90.61 & 84.35 \\
    \textbf{2} & 96.86 & 89.96 & 98.78 & 89.51 & 92.37 & 91.90 \\
    \textbf{3} & 96.88 & 89.62 & 98.95 & 90.28 & 92.72 & 92.36 \\
    \textbf{4} & 96.86 & 89.67 & 98.94 & 90.39 & 92.81 & 92.42 \\
    \bottomrule
    \end{tabular}
}
\label{tab::appendix:step_para}
\end{table}

\begin{table}[H]
\centering
\setlength{\tabcolsep}{6pt}
\setlength{\abovecaptionskip}{2pt}
\setlength{\belowcaptionskip}{2pt}
\caption{The influence of dilated depth $D$.}
\resizebox{0.9\linewidth}{!}{
    \begin{tabular}{c|cccccc}
    \toprule
    $D$ & \textbf{Enron} & \textbf{DBLP} & \textbf{AS733} & \textbf{FB} & \textbf{HepPh} & \textbf{MovieLens} \\
    \midrule
    \textbf{1} & 91.14 & 88.67 & 94.98 & 73.69 & 87.38 & 80.44 \\
    \textbf{2} & 92.16 & 89.89 & 97.12 & 83.18 & 90.41 & 87.52 \\
    \textbf{3} & 96.86 & 89.96 & 98.78 & 89.51 & 92.37 & 91.90 \\
    \textbf{4} & 97.17 & 89.62 & 98.93 & 89.99 & 93.17 & 91.47 \\
    \bottomrule
    \end{tabular}
}
\label{tab::appendix:depth_para}
\end{table}

\end{document}